\documentclass{article}

\usepackage{microtype}
\usepackage{graphicx}
\usepackage{subfigure}
\usepackage{booktabs} 

\usepackage{hyperref}

\usepackage{multirow}

\usepackage{amsmath}
\usepackage{amssymb}
\usepackage{mathtools}
\usepackage{amsthm}
\usepackage{graphicx}
\usepackage{subcaption}  
\usepackage{float} 
\usepackage[capitalize,noabbrev]{cleveref}
\theoremstyle{plain}
\newtheorem{theorem}{Theorem}[section]
\newtheorem{proposition}[theorem]{Proposition}
\newtheorem{lemma}[theorem]{Lemma}

\theoremstyle{definition}
\newtheorem{definition}[theorem]{Definition}

\theoremstyle{remark}

\usepackage[textsize=tiny]{todonotes}

\def\BibTeX{{\rm B\kern-.05em{\sc i\kern-.025em b}\kern-.08em
    T\kern-.1667em\lower.7ex\hbox{E}\kern-.125emX}}

\usepackage{soul} 

\usepackage{amsfonts}
\usepackage[linesnumbered,ruled,vlined]{algorithm2e}

\newcommand{\system}{\texttt{PCC}\xspace}
\newcommand{\model}{\texttt{SBD}\xspace}

\newcommand{\family}{\texttt{GSH}\xspace}
\newcommand{\find}{\texttt{SDPF}\xspace}
\newcommand{\snr}{\texttt{SNR}\xspace}
\newcommand{\mlc}{\texttt{MLC}\xspace}

\usepackage{tcolorbox}
\usepackage{marginnote}
\usepackage{bm}

\newcommand{\inputSeq}{\ensuremath{X}}     
\newcommand{\outputSeq}{\ensuremath{Y}}    
\newcommand{\sysBehavior}{\ensuremath{S}}  
\newcommand{\modelProb}{\ensuremath{p}}    
\newcommand{\variationalProb}{\ensuremath{q}} 

 \usepackage[preprint]{neurips_2026}

\usepackage[utf8]{inputenc} 
\usepackage[T1]{fontenc}    
\usepackage{hyperref}       
\usepackage{url}            
\usepackage{booktabs}       
\usepackage{amsfonts}       
\usepackage{nicefrac}       
\usepackage{microtype}      
\usepackage{xcolor}         

\title{To Explore The Strange New World Beyond Data Distribution: System Behavior, Causality Tax, and Non-causal Base Model}

\author{%
Xianzhi Zeng \\
 College of Computing and Data Science\\
  Nanyang Technological University\\
  Singapore, 639798 \\
  \texttt{xianzhi001@e.ntu.edu.sg} \\
  \And
  Jiangneng Li \\
  College of Computing and Data Science\\
  Nanyang Technological University\\
  Singapore, 639798 \\
  \texttt{JIANGNEN002@e.ntu.edu.sg} \\
\And
  Gao Cong \\
  College of Computing and Data Science\\
  Nanyang Technological University\\
  Singapore, 639798 \\
  \texttt{gaocong@ntu.edu.sg}
}

\begin{document}

\maketitle

\begin{abstract}

We show that the causality \footnote{Throughout this paper, we use \emph{causality} and \emph{causal formulation} interchangeably, referring to the causal factorization underlying autoregressive language models. They are not related to causal inference tasks involving interventions or counterfactuals here.} of language models (LMs) 
may not be necessary nor optimal. 
This is the case when \textbf{system behavior} (denoted as $\sysBehavior$)  
is incorporated as a first-principle Bayesian feature. 
{Here, $\sysBehavior$ refers to \textbf{extra dominant factors} beyond the data space, and they involve coupled effects.}
Despite being the de facto foundation of modern architecture, recent studies indicate persistent mismatches and contradictions with causality. These issues largely stem from system behavior rather than the data distribution. 
We therefore propose the \model framework, which incorporates $\sysBehavior$ as an irreducible component of the evidence lower bound (ELBO). {\model theoretically reveals a counter-intuitive \textbf{Causality Tax} phenomenon, where
causality emerges as a suboptimal approximation with an additional structural error, due to the obliviousness to $\sysBehavior$.}
{To address the challenge of latent variable analysis, we validate the \model-predicted impact of $\sysBehavior$ via implicit measurements, theoretical-bound-guided controls, and Neural Tangent Kernel (NTK) evaluations.}
%
%
In particular, we construct Green Shell (\family) to show the possibility of reducing Causality Tax. \family is a non-causal variational family, and it replaces the sequential dependency chain of  $\sysBehavior$ components with a divide-and-conquer partition. NTK spectra in the lazy-training regime confirm that \family always achieves significantly tighter error bounds than causality, 
with $7dB+$ improvement in signal-to-noise ratio.
In the relatively later stage of lazy-training, \family further 
 leads to superior generalization (up to $20\%$ richer multi-scale fitting capabilities).
Taken together, \model establishes system behavior as a complementary theoretical abstraction besides causality and distribution fitting, opening new research avenues such as designing and optimizing LM base models.


\end{abstract}

\section{Introduction}
\label{sec:intro}
\begin{figure}[t]
\includegraphics[width=\textwidth]{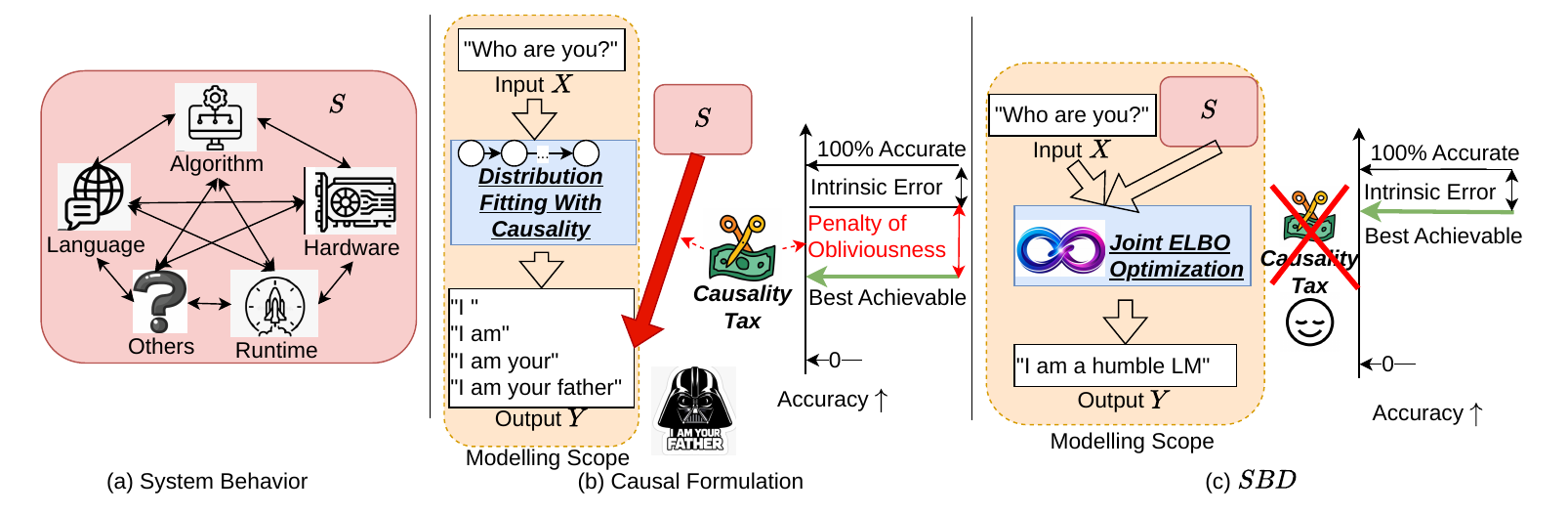}
\caption{Overview of \model.
}
\label{fig:overview}
\end{figure}

Modern autoregressive language models (LMs)~\citep{vaswani2017attention}, particularly large language models (LLMs)~\citep{yang2025qwen3,touvron2023llama,achiam2023gpt}, adhere to the \emph{causal formulation}: each token is generated conditioned on the incrementally growing prefix. 
{
However, we observe increasing mismatches and contradictions with causality from multiple aspects~\citep{atil2024non,pmlr-v235-papadopoulos24a,arriola2025block,fu2024break}. 
For instance, causality seems to conflict with Shannon's law~\citep{pmlr-v235-papadopoulos24a}, and a deliberate relaxation of causality is self-consistent and feasible~\citep{fu2024break,arriola2025block}.
Those contradictions originate systematically and typically beyond data distributions, and we therefore summarize these factors as \emph{system behavior} $\sysBehavior$ (Definition~\ref{def:system_behavior}). 
{
Theoretically, 
 $\sysBehavior$ is \emph{deliberately broad to cover all that is orthogonal to the data space}, as any narrower definition would arbitrarily exclude aforementioned systematic effects. In practice, even under the computational and observability constraints, the observable and manipulable subset $\sysBehavior$ is still far beyond the passive infrastructure of hardware and runtime~\citep{atil2024non}, such as those from algorithms, natural languages and other yet-to-be-understood factors~\citep{pmlr-v235-papadopoulos24a,fu2024break}.
}
}
\begin{definition}[System Behavior]
\label{def:system_behavior} In this paper, 
system behavior ($\sysBehavior$) refers to the union of all transformation factors beyond the data distribution in LMs, including algorithms, hardware, runtime, natural languages, and other yet-to-be-understood factors (Figure~\ref{fig:overview}(a)).
\end{definition}

%

Causality focuses on distribution alignment via a sequential dependency chain, e.g., ``I'' $\rightarrow$ ``I am'' in  Figure~\ref{fig:overview}(b). However, it remain \textit{oblivious} to how $\sysBehavior$ governs the generation process (red arrows in Figure~\ref{fig:overview}(b)).
Intuitively, obliviousness leads to a penalty, such as inaccurate or even offensive outputs~\citep{ouyang2025empirical,betley2025emergent} (e.g., ``I am your father'').
Formally, causality may degrade the best achievable accuracy. The resulting error can exceed the intrinsic approximation error of the neural architecture due to the aforementioned mismatches.
%
Therefore, a fundamental question arises as:
\begin{center}
    \textbf{\emph{How can language modelling explicitly incorporate $\sysBehavior$ for the reconciliation, i.e.,  to identify, explain, and reduce these systematic penalties?}}
\end{center}

{Incorporating an often-overlooked space is usually harder than improving a well-established one. The formulation needs to be broad enough to give all factors beyond data a meaningful and interpretable role. It also needs to stay concrete enough to guide practical improvements, rather than becoming overly abstract.}
%
We propose the \textbf{S}ystem \textbf{B}ehavior \textbf{D}ynamics (\model), a Bayesian framework to bridge the gap. 
Bayesian framework~\citep{kingma2013auto} introduces \emph{latent variables} to capture the broad diversity without requiring rigid, manual specifications, and it utilizes the \emph{Evidence Lower Bound (ELBO)} to be compatible with gradient-based optimization.
This is different from point-estimates or functional constraints~\citep{bertsekas2014constrained}, which suffer from mode collapse or optimization instability when faced with complex system behaviors considered in this work.
We therefore treat $\sysBehavior$ as \emph{a set of latent variables}, and it is an irreducible component of the ELBO  (Figure~\ref{fig:overview}(c)). Such ELBO jointly considers the global consistency of $\sysBehavior$ and output (e.g., ``I am a humble LM''), and an expansion of it confirms that \textbf{\emph{there is a structural overhead, which we refer to as Causality Tax, to enforce the causal chain itself.}}
Furthermore, the causality tax is not inevitable. As shown by \model, the ELBO remains globally consistent even without imposing a causal chain.

%
%
{Despite the theoretical insights, validation and utilization of \model in practice is still challenging, due to the less explicit nature of investigating latent variables compared with fitting data distributions~\citep{blei2017variational}.
 Specifically, a closed-loop latent variable analysis~\citep{bollen2014structural} involves a technical question of 
\textbf{\emph{how to measure, control, and evaluate the impacts of $\sysBehavior$ within \model?}}}

%
We address these detailed aspects in turn.
{
First, we conduct a perturbation analysis for \emph{measurement}.
\model establishes that $\sysBehavior$ structurally shapes the output during inference beyond hardware or runtime randomness. 
Although some components of $\sysBehavior$ are unknown or unobservable, its structural influence arises jointly from algorithms, hardware, runtime environments, and natural languages. Therefore, we inject Gaussian random noise into input embeddings, and examine the resulting changes in output logits and hidden states. This analysis
reveals stable regularities in logit variability and hidden-state distances beyond simple noise propagation.
}
 %
{Second, we \emph{control} the grouping strategies of $\sysBehavior$ to especially show that the Causality Tax can indeed be reduced or even avoided.
%
We introduce Green Shell (\family), a non-causal variational family inspired by Green’s functions~\citep{green1889essay}, to replace causal formulations. 
\family achieves a tighter theoretical error bound than its causality counterparts by a spatial partition over $\sysBehavior$ components.
Instead of imposing a causal chain, such a spatial partition inherits a divide-and-conquer approximation from Green’s functions and thus gets rid of the sequential dependencies.
}
Finally, Neural Tangent Kernel (NTK) spectra in the lazy-training regime~\citep{jacot2018neural} offer a rigorous framework for model analysis, 
and we empirically validate causality-tax-reduction claims of \family through NTK \emph{evaluation}.
{Our evaluation also suggests that \family may support the development of new base model families. In particular, its decoupled structure and reorganization of $\sysBehavior$ suggest the potential for training language models without backpropagation.}



\textbf{Contributions}. Our main contributions are as follows:
\begin{itemize}
    \item We propose \model, a new abstraction of $\sysBehavior$ and treat it as a first-principle factor. (Section~\ref{sec:model}).
    \item We {discover the Causality Tax by utilizing \model}, indicating that causality is neither necessary nor optimal. (Section~\ref{sec:attack}).
    \item  We establish the closed-loop validation of \model beyond the theoretical claims through measurements, controls and evaluations around $\sysBehavior$. {A proof-of-concept non-causal variational family \family is proposed and examined during this establishment.}(Section~\ref{sec:validation}). 
\end{itemize}

\section{Motivation and Related Work}
\label{sec:related}
{This section identifies a growing mismatch with causality through a literature review, followed by relevant Bayesian concepts.
Discussions of more related work are shown in Appendix~\ref{sec:appendix_other_related}.}

\textbf{Indication of Mismatches}.
\label{subsec:mismatches}
Recent studies have indicated various mismatches with the causal formulation in autoregressive (AR) LMs. 
Empirically, non-determinism~\cite{atil2024non} is involved in inference, where identical inputs can lead to different outputs and fluctuating internal states. This phenomenon involves future states modifying past representations, {while causality enforces the one-way determinism of the past to shape the future}.
%
Theoretically, the Arrows of Time (AoT) framework~\cite{pmlr-v235-papadopoulos24a} uncovers a fundamental paradox between causality and the symmetry of information~\cite{shannon1951prediction}. Meanwhile, block diffusion~\cite{arriola2025block} demonstrates that the token-by-token causal chain itself can be reconstructed and shortened via a denoising process.
On the practical side, inference optimizations such as speculative decoding~\cite{leviathan2023fast} and lookahead decoding~\cite{fu2024break} deliberately relax strict token-by-token causality to improve efficiency, while still attempting to preserve the original probabilistic distribution.

Despite these valuable indications of mismatch, causality remains the foundational assumption of all the above works. In contrast, our work revisits the foundation of causality from first principles. 
\textbf{Bayesian Modelling, Variational Inference, and Evidence Lower Bound}.
{
Beyond language modelling, there is a growing trend to treat the \emph{system itself} as an active component of the behavioural model. Representative examples include tiered replication for failure-aware placement~\cite{cidon2015tiered}, attention-aware speculative prefetching in InfiniGen~\cite{lee2024infinigen}, and optimizing streaming database system operators in PECJ~\cite{zeng2024pecj}. Our work is the first to extend this modelling philosophy to language modelling.
}

{
While cost analysis~\cite{lee2024infinigen} and Markov chains~\cite{cidon2015tiered} also provide valuable insights for system optimization, Bayesian modelling, as involved in PECJ~\cite{zeng2024pecj}, offers a unified treatment of uncertainty. Instead of modelling each source of randomness individually and passively reacting, PECJ proactively compensates for oscillation by abstracting both observable and unobservable factors into latent variables and formulates a Posterior Distribution Approximation (PDA) problem. This is regardless of database deployment details such as hardware placement, network conditions, or request skew. Variational Inference (VI)~\cite{bbvi_2014,svi_2015} is then utilized to make PDA tractable by approximating the posterior with a variational family and optimizing the Evidence Lower Bound (ELBO). For problems of moderate complexity, such as stream oscillation compensation in PECJ, both analytical methods (e.g., coordinate ascent variational inference ~\cite{cavi}) and black-box approaches (e.g., BBVI~\cite{bbvi_2014}) are applicable to optimize ELBO. As the problem complexity increases, such as image, text or video generation~\cite{kingma2013auto,arriola2025block}, closed-form coordinate updates becomes increasingly difficult, and BBVI provides a more practical alternative.
}

{
It is worth emphasizing that, although ELBO is also widely used in supervised learning frameworks such as variational autoencoders~\cite{kingma2013auto}, it should not be interpreted through the lens of log-likelihood optimization. The theoretical foundation of ELBO is Bayesian evidence rather than likelihood maximization. Accordingly, across most Bayesian formulations, including variational autoencoder, PECJ, and this paper, only information that is known \emph{a priori} and free from the system's intrinsic stochasticity serves as evidence for conditioning.
While the modelling objectives differ substantially across applications, this Bayesian treatment of evidence remains unchanged.
}

\section{{System Behavior in Bayesian Modelling}}
\label{sec:model}
In this section, we formulate \model from a Bayesian statistics aspect. {We first discuss the key differences regarding abstraction principles, then discuss the detailed meanings.  
}
\subsection{A Shift of Abstraction: Ascending the Latent Variables}
\textbf{Causal Formulation: Latent Variables as Slaves of Distribution Fitting}.
Causality is tightly linked with the data distribution. Even in relaxed frameworks such as block diffusion, this binding persists.
Formally, both autoregressive models and block diffusion optimize the ELBO through sequential or structured causal generation. For block diffusion, this involves a latent corruption process \( q(Z \mid X) \) and a reverse generation model \( p_\theta(X \mid Z) \), yielding Eq.~\ref{eqn:elbo_tradition}, where \( Z \) denotes latent variables of the causal chain in the corrupted space. This formulation also subsumes autoregressive generation as a special case when the corruption process is instantiated as successive token masking~\cite{austin2021structured}.
\begin{equation}
    \log p_\theta(X) \geq \mathbb{E}_{q(Z \mid X)} \left[ \log p_\theta(X \mid Z) \right] - \mathrm{KL}(q(Z \mid X) \| p(Z)), 
    \label{eqn:elbo_tradition}
\end{equation}

Such data-driven formulations treat observed data \( X \) as fixed external evidence and evaluate quality primarily by reconstruction fidelity. However, many contradictions to causality discussed before arise \textbf{not} from the data distribution.
{For instance, non-determinism comes from \emph{runtime} and \emph{hardware} behaviors~\cite{atil2024non}, AoT is due to several \emph{algorithm} aspects, including computability, irreversibility of model architectures and operator implementations~\cite{pmlr-v235-papadopoulos24a}, and look-ahead decoding relies on an alternative \emph{algorithm} paradigm of forwarding an LM~\cite{fu2024break}.}
{
Besides, there are system-wide behaviors that mainly arise from the \emph{language} itself and \emph{potentially unknown factors}, also beyond a pure data distribution, such as reasoning collapses under complexity shifts~\cite{illusion-of-thinking}, the misalignment caused by narrow finetune~\cite{betley2025emergent},  and the emergence of metacognitive self-monitoring~\cite{didolkarmetacognitive} (Appendix~\ref{subsec:cog_discuss}).}

\textbf{To Learn From PECJ: Latent Variable Usage $\neq$ Distribution Fitting.}
The mathematical formalism of PECJ directly inspired our modelling and derivations. 
{
In PECJ, there is also a latent variable notation \( Z \). Different from Eq.~\ref{eqn:elbo_tradition}, it captures various environmental factors that induce uncertainty in the streaming database engine, such as network transmission delay and OS scheduling jitter. In other words, this is an orthogonal factor beyond data. However, the key objective of PECJ is to proactively compensate for the so-called stream oscillations in the window join algorithm, and $Z$ is eventually marginalized by a set of BBVI techniques~\cite{bbvi_2014} during achieving this goal. While this kind of marginialization process may lead to a more comprehensive loss function compared with Eq.~\ref{eqn:elbo_tradition}, it is still not clear how and why these factors affect.
}


\textbf{To Form the \model Abstraction: Latent Variable $\sysBehavior$ as the First Principle.}
Following PECJ, we reformulate the ELBO to characterize the global consistency rather than merely distribution fitting quality. However, instead of marginalizing latent variables, we consider $\sysBehavior$ as the first principle.
In this view, accurately modelling language generation requires accounting for three elements simultaneously: the input data \(\inputSeq\), the output data \(\outputSeq\), and the system behavior \(\sysBehavior\). %
{
Specifically, given a joint distribution $\modelProb(\outputSeq,\sysBehavior,\inputSeq)$, there is a variational distribution $\variationalProb(\outputSeq,\sysBehavior)$ to approximate the posterior distribution $\modelProb(\outputSeq,\sysBehavior|\inputSeq)$.
The intuition is that a system itself should be self-consistent, rather than just being a good approximation mechanism of data input and output.
}
%
Taking the prior \(\log \modelProb(\inputSeq)\) as the overall evidence, this joint perspective leads to a system-behavior-aware ELBO as formalized in Eq.~\ref{eqn:elbo_paradigm}, and further motivates Proposition~\ref{pro:decouple_elbo}.

\begin{equation}
\text{ELBO}_{\variationalProb} = \mathbb{E}_{\variationalProb(\outputSeq, \sysBehavior)} \left[ \log \modelProb(\outputSeq, \sysBehavior, \inputSeq) \right] - \mathbb{E}_{\variationalProb(\outputSeq, \sysBehavior)} \left[ \log \variationalProb(\outputSeq, \sysBehavior) \right] \label{eqn:elbo_paradigm}
\end{equation}

\begin{proposition}[Decoupled ELBO]\label{pro:decouple_elbo}
By expanding the $\modelProb(\outputSeq, \sysBehavior, \inputSeq)$ in Eq.~\ref{eqn:elbo_paradigm}, a further decoupled ELBO is derived as Eq.~\ref{eqn:elbo_final}. The $ \mathbb{E}_{\variationalProb(\outputSeq, \sysBehavior)}$ notation is removed for $\modelProb(\inputSeq)$, since there is no $\variationalProb$ involved. 
\begin{align}
        \text{ELBO}_{\variationalProb}
        &= \log \modelProb(\inputSeq)
          + \mathbb{E}_{\variationalProb(\outputSeq, \sysBehavior)}\left[\log \modelProb(\outputSeq \mid \inputSeq)   \right] 
      + \mathbb{E}_{\variationalProb(\outputSeq, \sysBehavior)} \left[
            \log \modelProb(\sysBehavior \mid \inputSeq, \outputSeq)
            - \log \variationalProb(\outputSeq, \sysBehavior)
        \right] \label{eqn:elbo_final}
        \end{align}
\end{proposition}

\subsection{Detailed Meanings}
We structurally discuss other terms in Proposition~\ref{pro:decouple_elbo} in the following.

\textbf{Token Alignment (TA)} \( \mathbb{E}_{\variationalProb(\outputSeq, \sysBehavior)}\left[\log \modelProb(\outputSeq \mid \inputSeq) \right] \).
During training, since the target \(\outputSeq\) is a fixed ground-truth, its variational distribution collapses to a Dirac delta, reducing the term into standard cross-entropy (CE, $\log \modelProb(\outputSeq \mid \inputSeq)$).
During inference, \(\outputSeq\) becomes stochastic, and the expectation must be retained to integrate over the latent space of \(\sysBehavior\).

More importantly, as a structural component of expectations, this term is expected to exhibit regular patterns beyond mere random noise. 
Since a well-trained language model has largely aligned its output distribution \(\outputSeq\) with the data during training, propagated deviations in the variational approximation during inference
is mostly with respect to \(\sysBehavior\) through this expectation. 
As a result, the TA term effectively encodes the joint influence of the input \( X \) and system behavior \(\sysBehavior\), causing the hidden states to exhibit systematic, non-random modifications induced by \(\sysBehavior\).
We leverage this property as a probe to implicitly measure \(\sysBehavior\) in Section~\ref{sec:measure}.

\textbf{System Response to Input/Output (SRIO)} \( \mathbb{E}_{\variationalProb(\outputSeq, \sysBehavior)} \left[ \log \modelProb(\sysBehavior \mid \inputSeq, \outputSeq) \right] \).
This term reflects the coherence between the system behavior \(\sysBehavior\) and the generated output \(\outputSeq\). It captures the model's inductive biases and expectations regarding its own structural dynamics, given a particular input-output pair. Such biases include convergence patterns, consistency of attention flows, and the plausibility of hidden state transitions.
For simplicity, in a frozen LM without external memory, we approximately treat this term as constant and work with its expectation.
We believe that explicitly expanding this term in settings involving memory mechanisms or agentic behaviors could significantly benefit subsequent optimization and interpretability of language models.

\textbf{Approximation's Internal Semantics (AIS)} \( -\mathbb{E}_{\variationalProb(\outputSeq, \sysBehavior)} \left[ \log \variationalProb(\outputSeq, \sysBehavior) \right] \). 
As  the entropy of the variational distribution itself,
AIS governs how the output generation process is internally organized and approximated within the system. 
{AIS is fixed during model inference, as ``how to conduct the approximation'' is not changed, given the frozen model weights.}
{On the other hand, model training is a BBVI optimization~\cite{bbvi_2014} over AIS.
Specifically, the approximation is implicitly optimized through changing model weights, and loss functions serve as the proxy to guide the optimizations and reduce approximation errors.}
{
For most BBVI implementations, including ours, different reasonable proxy objectives induce the same underlying error bound, regardless of the specific choice of loss function. This error bound is fundamentally determined by the structural organization of \(q(\cdot)\), which we further discuss in Section~\ref{sec:attack}. Common loss functions, such as cross-entropy, have already provided reasonable proxy objectives, but are not the only viable choices. More details are provided in Appendix~\ref{subsec:green_approx_appendix}.
}

\section{Discovery of the Causality Tax}
In this Section, we rethink causality and inverstigate further on the approximation error bound through two derived Lemmas.
\label{sec:attack}
%

\subsection{Causality is Not Necessary}
\label{subsec:causality_not_necessary}
AIS itself contains no explicit left-to-right generation order or fixed autoregressive structure (Proposition~\ref{pro:decouple_elbo}), and causality only re-emerges as a byproduct of specific structural assumptions.
A canonical example is diffusion models, where system behavior \(\sysBehavior\) is decomposed into a thermal-noising trajectory \(\sysBehavior = \{s^0, s^1, \ldots, s^T\}\), and the AIS takes the factorized form shown in Eq.~\ref{eqn:diffusion_cas}.
\begin{align}
    q(Y,S) & = \variationalProb(Y|s^0, s^1, \ldots, s^T) \times  \prod_{\text{For each step } i \neq 0} \variationalProb(s^i \mid s^{i-1}) \times \variationalProb(s^0)  \label{eqn:diffusion_cas}
\end{align} 

In this case,   
each \( s^i \) represents the system state at the \( i \)-th denoising step dependent on the previous step $i-1$ by definition. 
  A similar phenomenon occurs in autoregressive models, which replace $\variationalProb(s^i \mid s^{i-1})$ with a full historical token dependence $\variationalProb(s^i \mid s^{i-1}, \ldots, s^0, X)$, and conducts the cumulative product by ``For each token $i$ ''.
%
While such factorization can be effective, they impose strong causal structural constraints on the approximation of \(\sysBehavior\). 
In contrast, by treating the AIS variational distribution \( q(\outputSeq, \sysBehavior) \) as a free design choice rather than a fixed factorization, we can achieve more flexible and potentially superior control of approximation error. 
As long as the joint system converges to a high-probability configuration, the approximation remains valid. 



\subsection{Causality is Not Optimal}
\label{subsec:suboptimal_design}

\begin{lemma}[Expressivity Bound by Monotonicity of the Supremum]\label{thm:elbo}
\begin{equation}\label{eq:strict}
    \max (\mathrm{ELBO}_{q \in \Omega_{\mathrm{gen}}} )
    \;\;\ge\;\; \max (\mathrm{ELBO}_{q \in \Omega_{\mathrm{causal}}}),
\end{equation}
\end{lemma}

\begin{lemma}[Expressivity Bound by UAT]\label{thm:uat}
\begin{equation}\label{eq:uat-non}
    |F_{\mathrm{gen}}(X) - f(X)| < \epsilon
\end{equation}
 \begin{equation}\label{eq:uat-causal}
   |F_{\mathrm{causal}}(X) - f(X)| 
   < \epsilon + \sum_{t=1}^T |t_t(X)|
\end{equation}

  ${F}_{\mathrm{gen}}$ always achieves smaller approximation error than ${F}_{\mathrm{causal}}$, with strict inequality if  $\exists  t_t(X)\neq 0$.
  \end{lemma}
\textbf{An Intuitive Lemma to Claim Suboptimal Bound}.
Let \( f(X) \) denote the target output distribution \( p(Y \mid X) \) given input \( X \). Let \( F_{\mathrm{non}}(X) \) and \( F_{\mathrm{gen}}(X) \) be neural approximation mechanisms from the general variational inference family \( \Omega_{\mathrm{gen}} \) and the causal family \( \Omega_{\mathrm{causal}} \), respectively.
 By the \emph{monotonicity of the supremum}~\cite{rudin2021principles}, we observe \( \Omega_{\mathrm{causal}} \) is not optimal. Since the causal variational family \(\Omega_{\mathrm{causal}}\) is a strict subset of the general family \(\Omega_{\mathrm{gen}}\) (i.e., \( \Omega_{\mathrm{causal}} \subsetneqq  \Omega_{\mathrm{gen}} \)), any upper bound on \(\{\mathcal{L}(q): q \in \Omega_{\mathrm{gen}}\}\) is also an upper bound on \(\{\mathcal{L}(q): q \in \Omega_{\mathrm{causal}}\}\).
By instantiating \(\mathcal{L}(q)\) as the KL divergence and substituting into Eq.~\ref{eqn:elbo_final}, Lemma~\ref{thm:elbo} follows immediately. 

\textbf{A Non-trival Lemma Revealing of \emph{Causality Tax}}.
{
Lemma~\ref{thm:elbo} is equivalent to Lemma~\ref{thm:uat}, which is less intuitive but derived by \{a set of functional integration and inequality scaling as follows}. 

For the general family, the Universal Approximation Theorem (UAT) directly yields a tight error bound as shown in Eq.~\ref{eq:uat-non}.
For the causal family, the conditional independence constraints is additional imposed, as formalized in Eq.~\ref{eq:appendix_indep}, introducing an additional residual term \( t_t(X) \) defined in Eq.~\ref{eq:appendix_tt}. 
This residual term quantifies the deviation caused by enforcing conditional independence at each step of the causal chain. 
Specifically, the integration of \(\sysBehavior\) (or equivalently over the decomposed internal states \( s \)) over $\inputSeq$ ensures that the causal factorization must hold \emph{for all possible realizations} of the input data distribution \( X \), regardless of its specific form. In other words, the model must satisfy the conditional independence not only on the observed data, but across the entire support of the input distribution.

To simultaneously satisfy these causal constraints while approximating the target distribution \( f(X) \), the approximation error under the same parameter capacity becomes
Eq.~\ref{eq:appendix_target-causal}. 
Applying the triangle inequality then yields the final UAT bound for the causal family in Eq.~\ref{eq:uat-causal}, which is strictly looser than the non-causal case by the accumulated residual terms \( \sum_{t=1}^T |t_t(X)| \). $T$ is the length of the causal chain. For diffusion models like block diffusion, $T$ equals to the number of denosing steps; for autoregressive models,  $T$ equals to the number of generated tokens.

\begin{equation}\label{eq:appendix_indep}
  \left\{
  \begin{aligned}
  &\textbf{Diffusion:} && s_t \perp \{s_0,\dots,s_{t-2}\}\;\big|\;(s_{t-1},X), \\[4pt]
  &\textbf{Autoregressive:} && s_t \perp \{s_{t+1},\dots,s_T\}\;\big|\;(s_{0:t-1},X)
  \end{aligned}
  \right.
  \end{equation}
  
  \begin{equation}
  \left\{
  \begin{aligned}
  &\textbf{Diffusion:} 
  t_t^{\text{(diff)}}(X) = \\ 
   &\!\int\! p(s_t,\{s_0,\dots,s_{t-2}\}\mid X) \\ 
   &- p(s_t\mid X)p(\{s_0,\dots,s_{t-2}\}\mid X)\, dX, \\[4pt]
  & \textbf{Autoregressive:} 
  t_t^{\text{(AR)}}(X) = \\ 
  & \!\int\! p(s_t,\{s_{t+1},\dots,s_T\}\mid X) \\ 
   &- p(s_t\mid X)p(\{s_{t+1},\dots,s_T\}\mid X)\, dX
  \end{aligned}
  \right.
  \label{eq:appendix_tt}
  \end{equation}

\begin{equation}\label{eq:appendix_target-causal}
  |F_{\mathrm{causal}}(X) -\sum_{t=1}^T t_t(X)- f(X)| 
  < \epsilon 
\end{equation}

{
Compared with Lemma~\ref{thm:elbo}, the non-trival derivations behind leads to a specific term of \emph{Causality Tax}, which reveals that \emph{the causal chain itself leads to the additional error}.
}
By Universal Approximation Theorem (UAT), the general family achieves the approximation error bound in Eq.~\ref{eq:uat-non}, which reflects its ability to approximate the target \( Y \)distribution directly.
For the causal family with equivalent parameters and imposed with $T$-order causal chain, however, the UAT bound becomes strictly looser due to an additional residual term $\sum_{t=1}^T |t_t(X)|$ (Eq.~\ref{eq:uat-causal}). As derived in Eq.~\ref{eq:appendix_tt}, this residual term arises from the extra effort required to enforce the causal chain. Furthermore, each $t_t(X)$ is a hierarchy integration over all possible realizations of the $\inputSeq$ distribution, and the summation puts the guarantee at each order $t$ of the causal chain together.
%

%

%

%


\section{Validation and Analysis of \model}
\label{sec:validation}


In this Section, we verify \model through measurements, controls, and evaluations.
\subsection{Measurement on the Impacts of $\sysBehavior$}
\label{sec:measure}
We perform a perturbation analysis for the measurement. We utilize LLM (i.e., at least 3B) here rather than a smaller LM, as a smaller LM may involve a similar pattern, but weaker to be observed.

\textbf{Perturbation Setup.}
We run standard autoregressive decoding on a frozen LM and record the logits drift sequence $E$ associated with input tokens \( X \) across \( N \) generated steps, and a similar tracing is conducted on the last hidden states, denoted as the $r$ sequence.
For simplicity, both \( e_i \) and \( r \) are scalars after summation. A perturbed input is then constructed by injecting Gaussian noise into the tail of the token embeddings of \( X \). After that, we compute the normalized cross-correlation between \( E \) and \( r \) (or \(1/r\)), the maximum correlation value is \(\rho^\star\) and the best lag is \( k^\star \).
Logits are linearly related to the final-layer hidden states. If $\sysBehavior$ is a passive artefact that merely propagates injected noises~\cite{atil2024non,song2024good,ouyang2025empirical}, the cross-correlation should also link more tightly with linearity. However, Proposition~\ref{pro:decouple_elbo} implies the opposite. Since the token alignment (TA) term jointly depends on both \( X \) and \(\sysBehavior\), these changes should exhibit regular, non-random patterns. 

{Intuitively, the perturbation analysis can be viewed as probing a pizza factory pipeline consisting of three manufacturing stages: ``Crusting $\rightarrow$ Topping $\rightarrow$ Baking,'' where padding noise is analogous to deliberately introducing spoiled caviar at the very beginning of the production line. If the pipeline is merely a passive production process (analogous to $\sysBehavior$ consisting only of hardware/runtime noise), the contaminated ingredients before Baking should simply produce proportionally worse pizzas, leading to a strong correlation of $E \propto r$. In contrast, if the production process contains hidden mechanisms that actively respond to the ingredients (analogous to the language- and algorithm-dependent dynamics within $\sysBehavior$), the resulting degradation will deviate from this simple proportional relationship. }

Formally, a perturbed input is then constructed by injecting Gaussian noise into the tail of the token embeddings of \( X \). We collect the final-layer hidden states under this perturbation (\( H^{\text{curr}} \)) and compare them with the clean target hidden states \( H^{\text{target}} \). The overall discrepancy is quantified by the scalar \( r \), defined as the \(\ell_2\) norm of the element-wise Euclidean distances between \( H^{\text{target}} \) and \( H^{\text{curr}} \).
For simplicity, both \( e_i \) and \( r \) are scalars after summation.
A measurement metrics is to compute the normalized cross-correlation between \( E \) and \( r \) (or \(1/r\)) as 
\(
\hat{R}_{E,r}(k) = \frac{\sum_{i=1}^{N} e_i \, r_{i+k}}{\sqrt{\sum_{i=1}^N e_i^2} \sqrt{\sum_{i=1}^N r_i^2}},
\)
where \( k \) is the lag. We then identify the maximum correlation value \(\rho^\star\) and its corresponding best lag \( k^\star = \arg\max_k \hat{R}_{E,r}(k) \).

\begin{figure}[t]
     \begin{minipage}[c]{0.73\textwidth}
     \subfigure[$\rho^\star$ of $E$ and $r$.]{
            \includegraphics[width=0.3\textwidth]{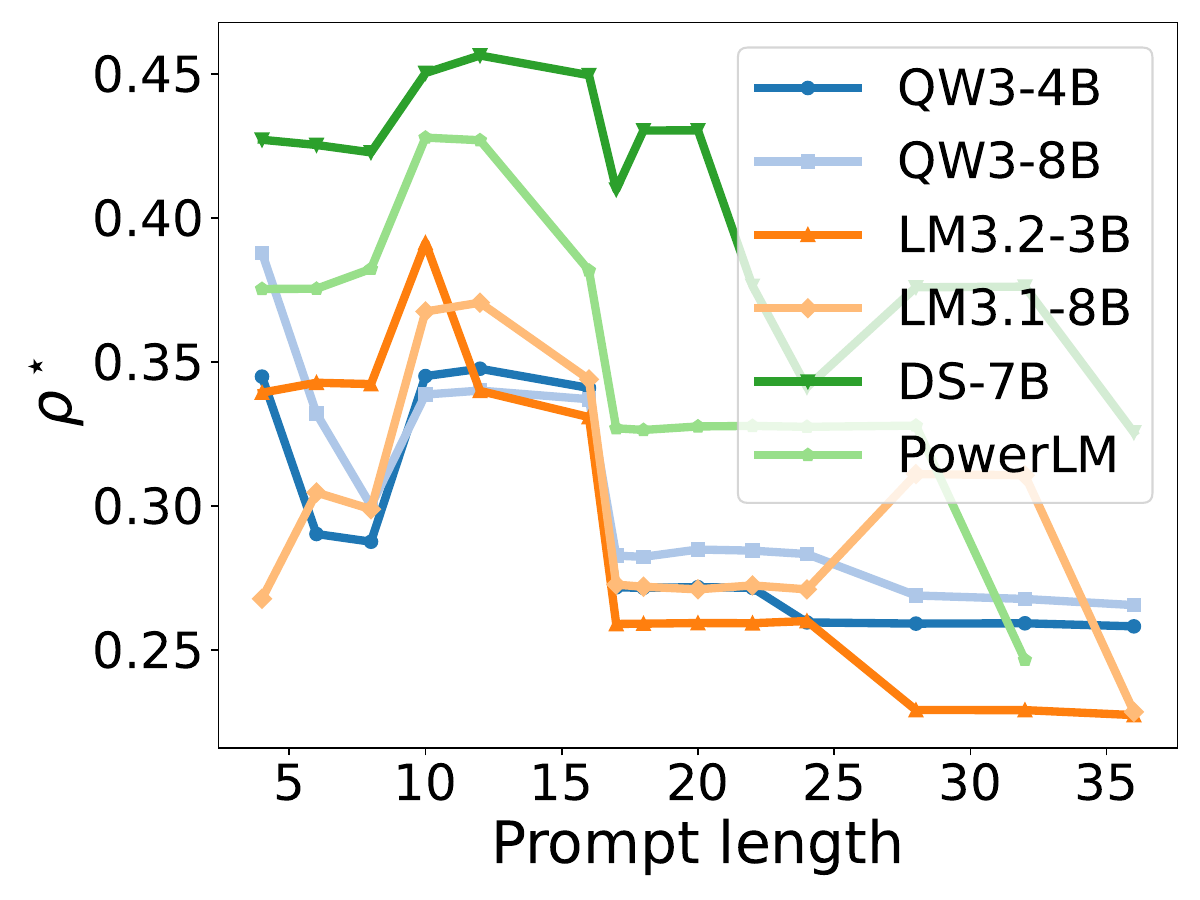}
            \label{fig:corr_1}
        }
        \subfigure[$\rho^\star$ of $E$ and $1/r$.]{
            \includegraphics[width=0.3\textwidth]{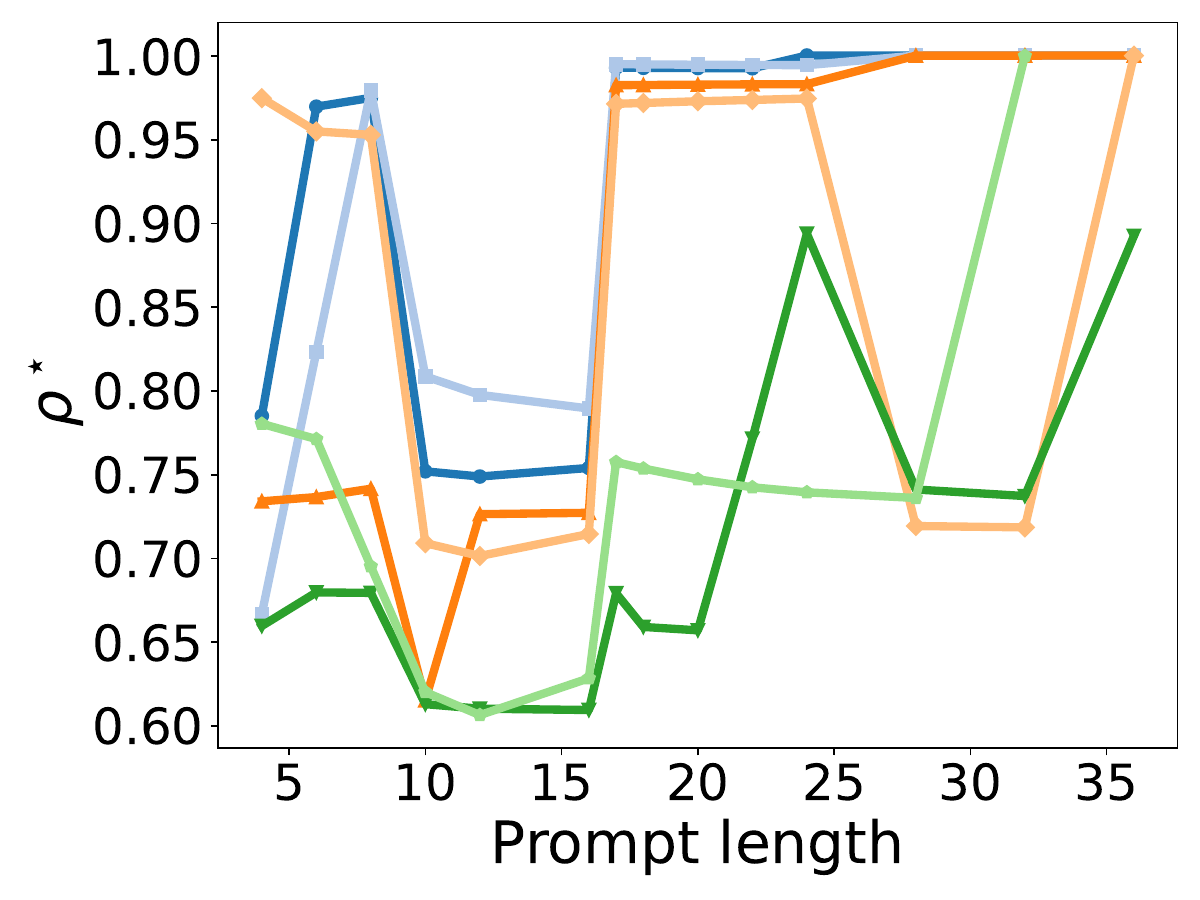}
            \label{fig:corr_2}
        } 
        \subfigure[ \(k^\star\) of $E$ and $1/r$.]{
            \includegraphics[width=0.3\textwidth]{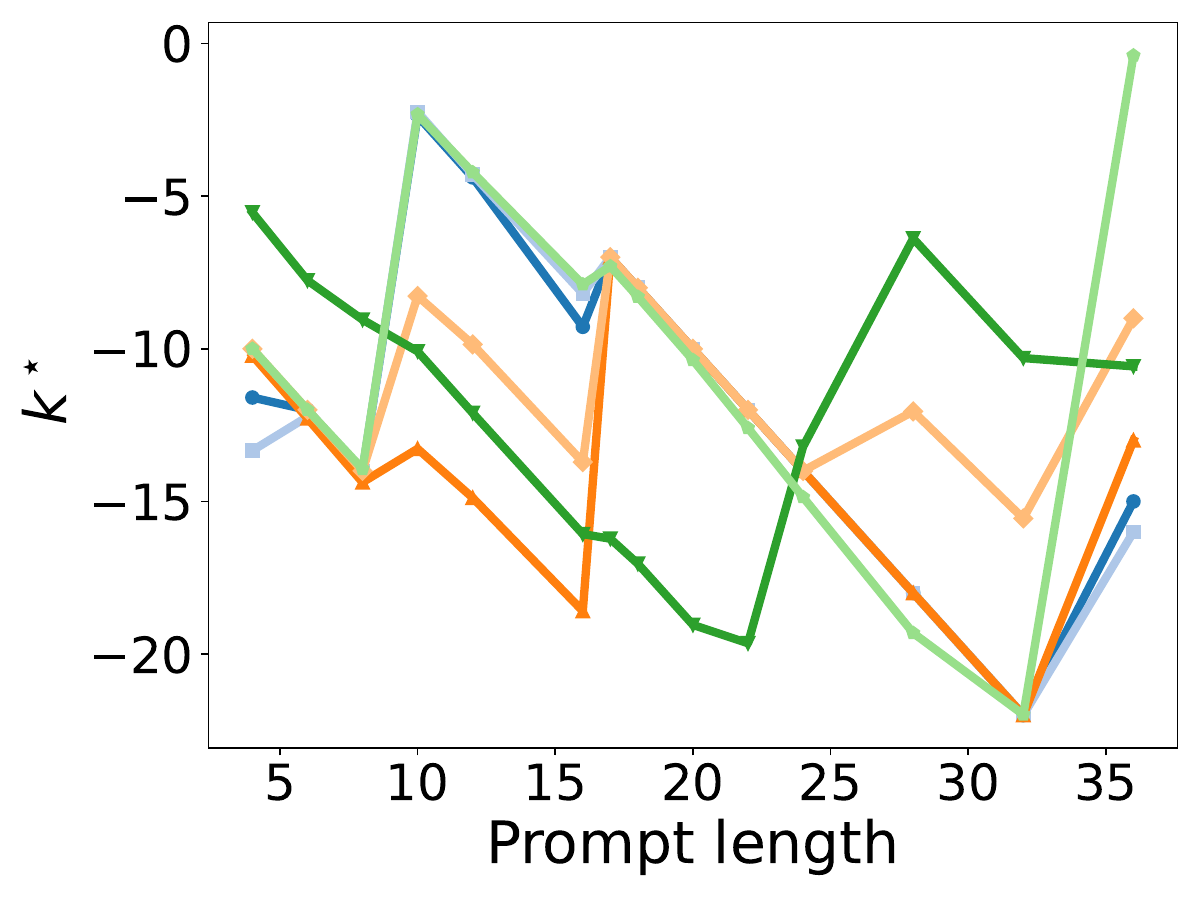}
            \label{fig:powerlaw_1}
        }
\caption{Implicit Measurement of $\sysBehavior$}
    \label{fig:cross_correlation}
    \end{minipage}
      \begin{minipage}[c]{0.25\textwidth} 
      \vspace*{-2mm}
        \includegraphics*[width=0.99\textwidth]{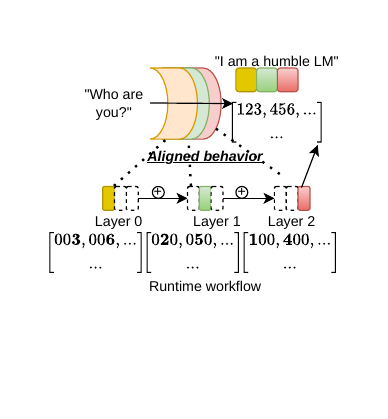}
        \vspace*{-7mm}
        \caption{\family.}
        \label{fig:gsh}
    \end{minipage}
  \end{figure}

\textbf{Measurement Results.} 
We let {QWen3-4B} (\emph{QW3-4B} for short)~\cite{yang2025qwen3}, {QWen3-8B} (\emph{QW3-8B} for short)~\cite{yang2025qwen3}, Llama3.2-3B~\cite{dubey2024llama} (\emph{LM3.2-3B} for short), Llama3.1-8B~\cite{dubey2024llama} (\emph{LM3.1-8B} for short), DeepSeek-7B~\cite{guo2025deepseek} (\emph{DS-7B} for short), and PowerLM-3B~\cite{shen2024power} to generate 24 output tokens, while varying the length of input tokens from $2\sim36$. We use random samples from \emph{TinyStories}~\cite{eldan2023tinystoriessmalllanguagemodels} 
to form the input, and report the results in Figure~\ref{fig:cross_correlation}. 
%
The probing results reveal clear systematic patterns. First, the correlation between \(E\) and \(1/r\) exceeds 0.6 for all models and reaches above 0.9 for some models with longer prompts, indicating a strong non-random relationship. On the other hand, the correlation with \(r\) is consistently lower (i.e., below $0.4$), indicating the linear propagation of random noise is weaker than other structural factors such as some inverse functional.
Furthermore, once the prompt length exceeds a certain threshold, the \(E\)-\(1/r\) correlation shows a sharp transition toward unity for \emph{QW3-4B}, \emph{QW3-8B} and \emph{LM3.1-8B}.
In addition,  \(k^\star\) typically deviates from \(-1\), indicating that TA term is not a simple instantaneous effect but involves a structured temporal propagation.
%

%

In summary, the fluctuations during inference exhibit measurable regularities beyond random noise. This arises from system behavior \(\sysBehavior\) that systematically influences token alignment (Proposition~\ref{pro:decouple_elbo}).
\subsection{Optimized Approximation Error about $\sysBehavior$}
\label{sec:so_what}
%

\textbf{Overview}.
To demonstrate that one can be free from the `Causality Tax’ discussed in Lemma~\ref{thm:uat}, 
we propose \family~(\textbf{G}reen \textbf{SH}ell), a non-causal mean-field variational family. 
{In physics, Green-Shell methods~\cite{jackson2021classical,green1889essay} are widely used, and 
the basic idea is to replace a complicated electromagnetic or potential field with a set of concentric shells as approximation units.
Instead of tracking every fine-grained interaction, one only needs to compute how charges or sources are distributed across these shells. 
In this way, Green-Shells approximate a complex system through spatial partitioning rather than temporal dependency, making each approximation unit inherently \emph{shared-nothing}.}
%
%
Inspired by these, \family
factorizes AIS as Eq.~\ref{eqn:green2} and 
partitions the  $\sysBehavior$ into $s^{i}$ by concentric shells instead of a causal chain. 
It still follows \(\sysBehavior = \{s^0, s^1, \ldots, s^T\}\), but $T$ here refers to the shell index.

{
Instead of instantiating a denoising chain  $\variationalProb(s^i \mid s^{i-1})$ (diffusion) or a full historical token dependence $\variationalProb(s^i \mid s^{i-1}, \ldots, s^0, X)$ (autoregressive), \family
instantiates $q(s^i)$ to align with field propagation in the Green-Shell structure, including:
(1) a temporal ordering constraint, requiring that the processing of shell $i$ occurs between shells $i-1$ and $i+1$ (the arrow in Figure~\ref{fig:gsh});  and
(2) a spatial partition constraint, requiring that shell $i$ occupies its own region of the hidden-state space (the layer partition at Figure~\ref{fig:gsh}, as concisely defined in Eq.~\ref{eqn:appendix_green1} and Appendix~\ref{sec:more_exmaple}). Given the spatial partition over $\sysBehavior$ and properties of Green-Shells, (3) the hidden-state space of $Y$ is also equivalently represented as spatial partitions $Y=y^0+y^1+...+y^T$, and each $y_i$ only covers the radius $R_{i-1}<r\leq R_{i}$ without overlap with others (partition of output in Figure~\ref{fig:gsh}).
}
%

{
Different from Eq.~\ref{eqn:diffusion_cas}, 
each shell $i$ performs an independent and unconstrained approximation of its local target $y^i$ through $\variationalProb(s^{i})$, without causal dependence on other shells $s^{j}$ (for $j \neq i$) or the input sequence $\inputSeq$. 
In this way, $\family$ incurs no additional structural approximation error beyond the intrinsic UAT bound $\epsilon$ (Eq.~\ref{eq:uat-non})~\cite{hornik1989multilayer,bishop2006pattern}, in the sense that (1) the temporal ordering constraint corresponds to a deterministic execution graph inside runtimes (e.g., PyTorch~\cite{Pytorch}); 
(2) the variational components $\variationalProb(s^{i})$ without causal chain dependency (e.g., $\variationalProb(s^{i}|s^{i-1})$ in Eq.~\ref{eqn:diffusion_cas}) are  \emph{sufficiently expressive in the infinite-width/depth regime of UAT}~\cite{hornik1989multilayer,bishop2006pattern}, thereby avoiding causal-chain induced terms $|t_t(X)|$ in Eq.~\ref{eq:uat-causal}); and (3) approximation error of $Y$ can be localized as $\variationalProb(y^i|s^i)$, rather than being always propagated through the coupled  $\variationalProb(Y|s^0, s^1, \ldots, s^T)$.
Please refer to Appendix~\ref{subsec:free_justification} for more details.
}

\begin{align}
    q(Y,S) & = \variationalProb(Y|s^0, s^1, \ldots, s^T) \times  \prod_{\text{For each layer } i} \variationalProb(s^{i})   = \prod_{\text{For each layer } i} (\variationalProb(s^{i}) \times \variationalProb(y^i|s^{i}))\label{eqn:green2}
\end{align} 

It is worth noting that the consistently directional field propagation preserves essential temporal and sequential information about \(\sysBehavior\) (as observed in Figure~\ref{fig:powerlaw_1}). 
Therefore, \family avoids both the collapse into a bag-of-words model and the rigid constraints of causality. 


\textbf{Loss Function and Parametrization.} A critical challenge in designing a loss function and parameterizing \family is that the mean-field of hidden-state and vocabulary tokens are \emph{not naturally aligned}. This leads to two main strategies: discretizing continuous fields into discrete tokens, or embedding 
discrete tokens into continuous field. {Both involve non-trival derivations as detailed in Appendix~\ref{subsec:green_approx_appendix}, and we briefly illustrate key insights in the following.}

In the discrete approach, the variational distribution becomes a categorical distribution defined by a nonnegative scoring function. Cross-entropy loss corresponds to one specific parametrization using exponential scoring and softmax normalization. Richer scoring schemes exist, such as label smoothing~\cite{szegedy2016rethinking}. Recent works~\cite{KalaiNachumVempalaZhang2025,peters2020sparse,tang2024learning} have also pointed out inherent limitations of cross entropy and suggested a broader design space beyond softmax.
In the continuous approach, the variational distribution is expressed by an energy functional, 
such as a Green Kernel (Eq.~\ref{eqn:appendix_kernel_layer} at Appendix~\ref{sec:more_exmaple}), and
a simpler quadratic energy leads to a mean-squared-error (MSE) loss. This formulation preserves the geometry of the field and offers numerical stability, but may discard richer structural dynamics. 
More sophisticated energy functions~\cite{lecun2006tutorial,zhang2023persistently} are left for future exploration. 

\textbf{Implementation and Limitations}.
\family is presented as a proof-of-concept rather than a fully engineered end-to-end base model. 
We briefly summarize the key implementation choices and limitations below. Please refer to Appendix~\ref{subsec:our_instantiation} for more details.
First, a complete removal of causality is incompatible with state-of-art infrastructures, such as standard KV caching mechanisms used in efficient inference. Therefore, we adopt a block-wise strategy similar to block diffusion~\cite{arriola2025block}, but replace the diffusion generation of each block with non-causal \family approximation.
Second, the parametrization of \family adopts a practical compromise. A continuous parametrization comes first, followed by a black-box renormalization under a discrete regime. Shell partitioning is currently performed via uniform radius division, and several auxiliary components (such as renormalization and diversity hyperparameters) are handled heuristically. 
\subsection{Evaluations about $\sysBehavior$ under NTK}
\label{sec:exp}

We evaluate on whether and how $\sysBehavior$ is optimized by \family under the lens of Neural Tangent Kernel (NTK)~\cite{jacot2018neural}, as well as examining the theoretical claims about error bounds at Section~\ref{sec:attack}. Please also refer to Appendix~\ref{sec:exp_additional} for additional sensitivity and ablation studies.

\textbf{NTK Proxy of Error Bound}.
Directly verifying the UAT error bound at Lemma~\ref{thm:uat} is extremely difficult in practice, and common end-to-end metrics (such as perplexity or accuracy) cannot be directly linked to the theoretical bound.
Fortunately, in the \emph{lazy training} regime, prior work has shown that the spectral properties of the NTK can be rigorously connected to generalization and approximation bounds~\cite{bartlett2002rademacher,caponnetto2007optimal,martin2021implicit}. However, this connection breaks down once the model enters the full feature-learning regime. We therefore conduct all evaluations in the lazy training regime to maintain theoretical validity.
Under this setting, we utilize the \textbf{Signal-to-Noise Ratio} (\snr) of the NTK spectrum as an approximately \emph{monotonic} proxy for the UAT error bound. 
Formerly, \snr is defined as the energy ratio of the \textit{signal subspace} to the \textit{noise bulk} in the NTK spectrum~\cite{martin2021implicit}, and a higher \snr corresponds to a tighter approximation bound (Appendix~\ref{sec:snr_detailes}).
Following standard practices in spectral analysis, we report \snr in decibels (dB), calculated as \( 10 \log_{10} () \).

In addition to \snr, we introduce another spectral metric called the \textbf{Main Lobe Concentration} (\mlc). For a given NTK spectrum, we define the main lobe as the set of eigenvalues consisting of the largest eigenvalue together with all subsequent eigenvalues whose energy is at least 20\% of the largest one. The \mlc is then computed as the ratio of the total energy contained in this main lobe to the total energy of the entire signal subspace.
A healthy NTK spectrum is expected to capture not only dominant features but also rich peripheral (edge) features. Therefore, a lower \mlc is generally preferable, as it indicates that the signal energy is more distributed across multiple significant eigenvalues rather than being overly concentrated in a few dominant modes.

\textbf{Other Setups}.
We conduct a controlled comparison on \family against two canonical baselines: 
(1) the classical \textbf{autoregressive (AR)} model~\cite{vaswani2017attention} 
and (2) the recent \textbf{block diffusion (BD)} model~\cite{arriola2025block}, which realizes a Markovian causal denoising process. 
Conceptually, there is a progressively relaxed causal hierarchy of $AR \rightarrow BD \rightarrow \family$. 
All models are trained from scratch under identical configurations and the same \textbf{150M-parameter backbone}. 
We utilize four datasets, i.e., \texttt{TinyStories}~\cite{eldan2023tinystoriessmalllanguagemodels},  \texttt{AG News}~\cite{ag_news_hf}, \texttt{MultiEURLEX}~\cite{chalkidis2021multieurlex} and \texttt{C4}~\cite{c4_hf}. 
{
Under identical experimental conditions in the lazy training regime, low-complexity datasets (e.g., TinyStories, MultiEURLEX) reach a relatively more advanced (``later stage'' for short) of convergence compared to semantically richer datasets (e.g., C4, AG News), which remain in an ``earlier stage''~\cite{rende2024distributional}
}.
Please refer to Appendix~\ref{sec:setups} for more details.
%

%

\begin{figure*}[t]
    \centering
  
    \begin{minipage}[c]{0.6\textwidth}
        \subfigure[C4]{
            \includegraphics[width=0.46\textwidth]{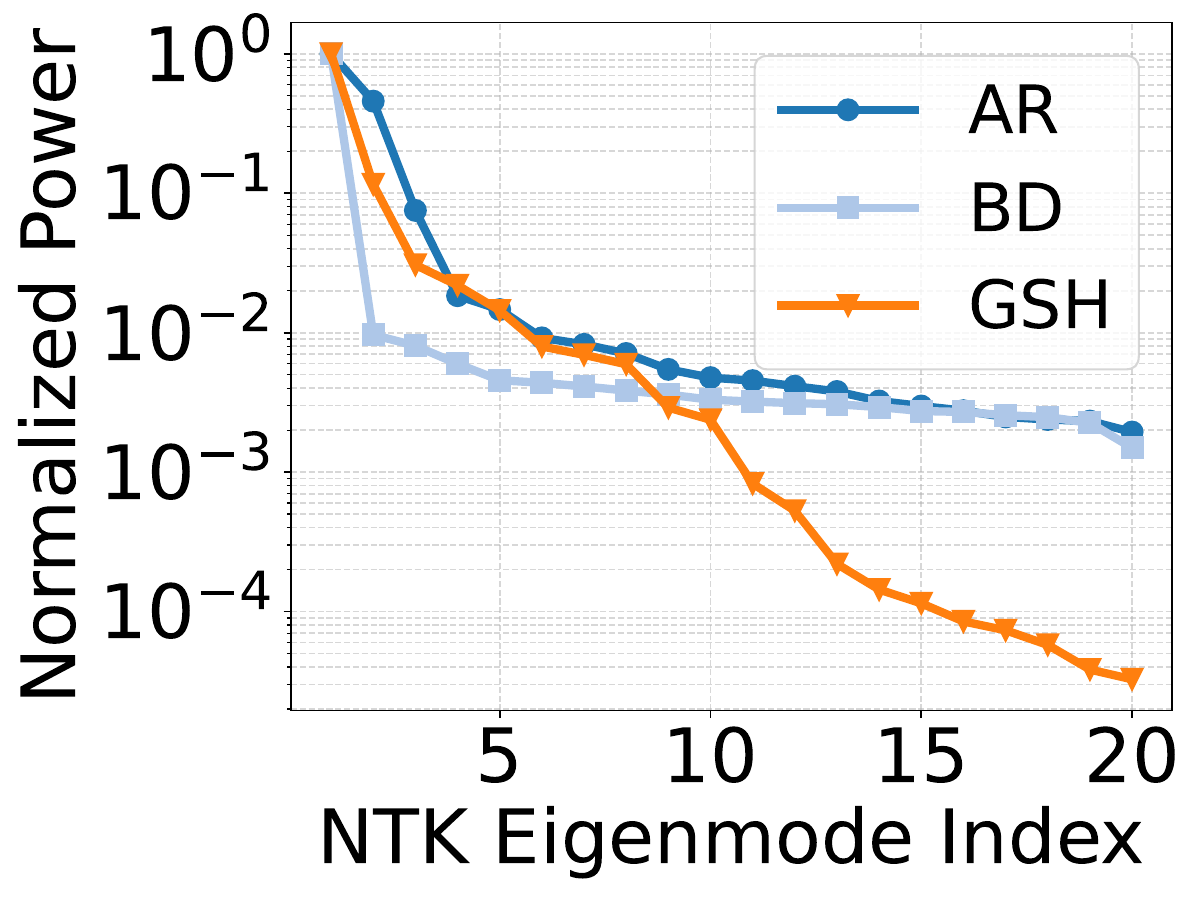}
        }
        \subfigure[TinyStories]{
            \includegraphics[width=0.46\textwidth]{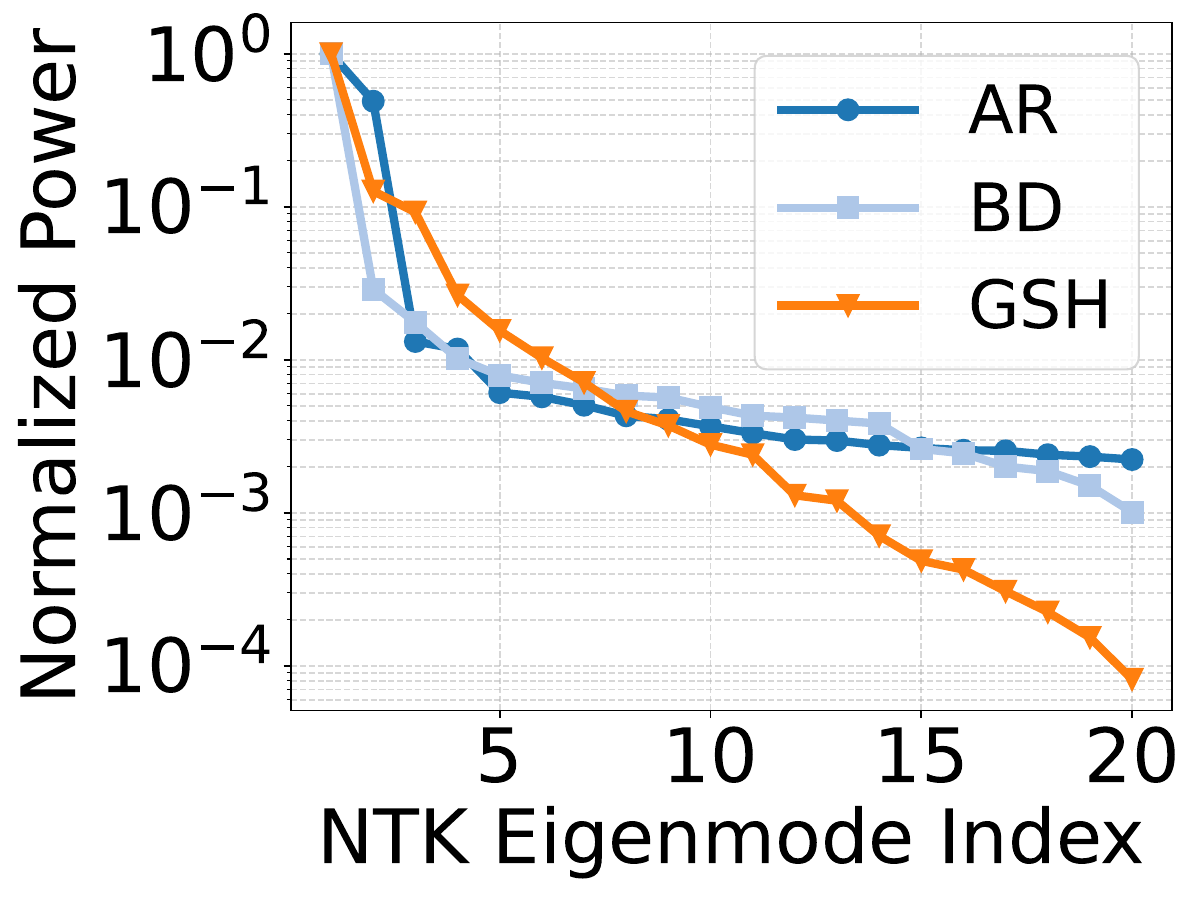}
        }
        \caption{Detailed NTK spectra.} 
    \label{fig:ntk_e2e}
    \end{minipage}
     \begin{minipage}[c]{0.3\textwidth}
        \includegraphics*[width=0.99\textwidth]{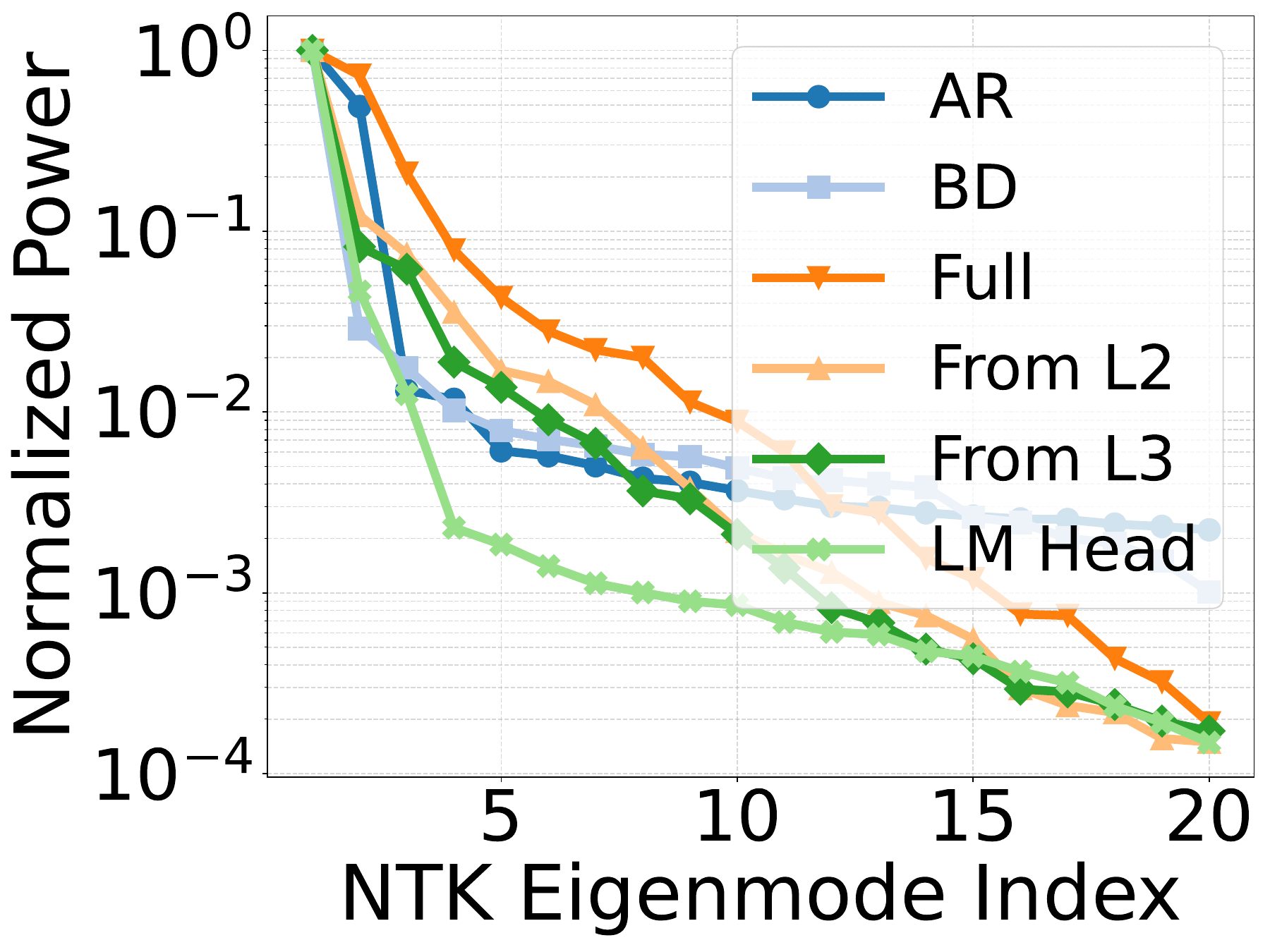}
        \caption{Resulting NTK spectra of gradient cutoff.}  
    \label{fig:bonus}
    \end{minipage}
  \end{figure*}

\paragraph{General Comparison.}
\begin{table}[]
\center
\caption{Overall Comparison.}
\label{tab:e2e}
\resizebox{.8\textwidth}{!}{
\begin{tabular}{|llll|llll|}
\hline
\multicolumn{4}{|l|}{Later Phase}                                                                                                         & \multicolumn{4}{l|}{Earlier Phase}                                                                                                   \\ \hline
\multicolumn{1}{|l|}{Dataset}                      & \multicolumn{1}{l|}{Method}  & \multicolumn{1}{l|}{\snr (dB, $\uparrow$)} & \mlc ($\downarrow$) & \multicolumn{1}{l|}{Dataset}                  & \multicolumn{1}{l|}{Method}  & \multicolumn{1}{l|}{\snr (dB, $\uparrow$)} & \mlc ($\downarrow$) \\ \hline
\multicolumn{1}{|l|}{\multirow{3}{*}{TinyStories}} & \multicolumn{1}{l|}{AR}      & \multicolumn{1}{l|}{12.8}                  & 0.94                & \multicolumn{1}{l|}{\multirow{3}{*}{AG News}} & \multicolumn{1}{l|}{AR}      & \multicolumn{1}{l|}{19.5}                  & 0.96                \\ \cline{2-4} \cline{6-8} 
\multicolumn{1}{|l|}{}                             & \multicolumn{1}{l|}{BD}      & \multicolumn{1}{l|}{15.8}                  & 0.92                & \multicolumn{1}{l|}{}                         & \multicolumn{1}{l|}{BD}      & \multicolumn{1}{l|}{15.6}                  & 0.85                \\ \cline{2-4} \cline{6-8} 
\multicolumn{1}{|l|}{}                             & \multicolumn{1}{l|}{\family} & \multicolumn{1}{l|}{\textbf{22.5}}         & \textbf{0.72}       & \multicolumn{1}{l|}{}                         & \multicolumn{1}{l|}{\family} & \multicolumn{1}{l|}{\textbf{28.3}}         & 0.95                \\ \hline
\multicolumn{1}{|l|}{\multirow{3}{*}{MultiEURLEX}} & \multicolumn{1}{l|}{AR}      & \multicolumn{1}{l|}{17.0}                  & 0.91                & \multicolumn{1}{l|}{\multirow{3}{*}{C4}}      & \multicolumn{1}{l|}{AR}      & \multicolumn{1}{l|}{17.3}                  & 0.93                \\ \cline{2-4} \cline{6-8} 
\multicolumn{1}{|l|}{}                             & \multicolumn{1}{l|}{BD}      & \multicolumn{1}{l|}{13.4}                  & 0.89                & \multicolumn{1}{l|}{}                         & \multicolumn{1}{l|}{BD}      & \multicolumn{1}{l|}{15.1}                  & 0.91                \\ \cline{2-4} \cline{6-8} 
\multicolumn{1}{|l|}{}                             & \multicolumn{1}{l|}{\family} & \multicolumn{1}{l|}{\textbf{24.0}}         & \textbf{0.73}       & \multicolumn{1}{l|}{}                         & \multicolumn{1}{l|}{\family} & \multicolumn{1}{l|}{\textbf{32.2}}         & 0.89                \\ \hline
\end{tabular}
}
\end{table}
The experimental results confirm our theoretical predictions. As shown in Table~\ref{tab:e2e}, \family consistently achieves the best  across all datasets. 
This provides direct empirical evidence that the non-causal variational family attains a tighter approximation error bound, as theoretically predicted in Section~\ref{sec:attack}.
In the earlier stage, \family starts with an extremely high \snr and highly concentrated main lobe, reflecting a conservative initialization. 
In the later stage, where the model has learned richer representations and covered more signals, \family not only obtains the highest \snr but also exhibits the lowest \mlc.
The detailed NTK spectra comparison at C4 and TinyStories (Figure~\ref{fig:ntk_e2e}) show that \family exhibits a ``lifting force'' during the transition from earlier to later. Such force compresses the main lobe and fills the mid-spectrum, leading to a more balanced signal distribution that suppresses noise without losing feature richness. Eventually, the signal energy is more effectively distributed across multiple significant eigenvalues. Therefore, besides the proven tighter error bound, non-casual variational family may also enjoy additional advantages in generalization capability.


\textbf{Bonus: Possibility of Backward Propagation Free.} 
There are 4 transformer layers labeled $0\sim3$ and one linear LM Head in our backbone NN, and we investigate the effect of gradient backward propagation cutoff from a certain layer in Figure~\ref{fig:bonus}. 
In this cutoff setting, the front part involves each layer training independently (i.e., only step 1 at Appendix~\ref{subsec:our_instantiation}), while the rest still conducts backward propagation across layers (i.e., both steps at Appendix~\ref{subsec:our_instantiation}). All other settings remain consistent with Figure~\ref{fig:ntk_e2e}-(b).
In general, \family outperforms state-of-the-art if the gradient cutoff occurs from the transformer layers, with stable \snr $\approx 22dB$ and \mlc $\approx 0.8$. The only significant downgrade occurs when the gradient cutoff starts at LM Head, and energy is too much concentrated at the first eigenvalue (\mlc=0.94). 
 Such possibilty of gradient cutoff is due to the layer decoupling nature of \family (Section~\ref{sec:so_what}), and further implies that backward-propagation-free training~\citep{li2025noprop} is not confined to small networks but is also achievable in larger LMs.
%
\section{Discussions}
\label{sec:discussion}
In this section, we discuss the limitations, potential risks of misuse and future work.
\subsection{Limitations and Impact Statement}
This work has inherent limitations as follows.
First, we identify the role of system behavior in \model with measurements, controls and evaluations, all around identifying and re-organizing latent variables $\sysBehavior$. The more detailed expansion of each $\sysBehavior$ component is left for future work.
Second, the Lemmas rely on the universal approximation theorem or the {monotonicity of the supremum}, and thus hold only within their validity regimes.
Third, the non-causal instantiation \family depends on field-theoretic approximations and black-box renormalization, leaving parts of its behavior analytically opaque.
%
%
Finally, our experiments are confined to the NTK spectra analysis and are intended to validate the theory rather than demonstrate end-to-end practical performance.

{
Given the first-principle changes of rethinking causality and system behavior, \model inevitably raises broader ethical and societal implications.  
}
In particular, malicious actors may deliberately and explicitly manipulate the system behavior $\sysBehavior$ to steer the coupled system of model weights, training vulnerabilities, or even infrastructure components, toward new attack planes, such as new risks of emergent misalignment~\cite{betley2025emergent}.
We discuss the details in Appendix~\ref{sec:risks}. 

\subsection{Future Work}
\textbf{Intermediate Next Step.}
{
One of the most promising avenues for future work is exploring the backward-propagation-free potential of \family (Figure~\ref{fig:bonus}). 
Beyond the showing the possibility of Causality-Tax-free, 
family directly aligns with more advanced LLM base models and the democratization of AI~\cite{UN_AI_Advisory_Body_2024}. 
While model training incurs a substantial device memory footprint to maintain the complete computational graph for gradients, non-causal variants like \family offer a principled mechanism to alleviate this overhead. 
Consequently, they not only yield tighter error bounds but also eliminate the need to retain the full gradient graph in device memory. 
While a comprehensive empirical exploration of this vision lies beyond the scope of the present work, active investigations are currently underway. 
In particular, further efforts are required to establish thorough benchmarks across multiple key dimensions. 
These include (1) knowledge and reasoning capabilities, (2) complex spatial-structural task metrics, (3) long-context processing, and (4) safety and robustness diagnostics~\cite{cao2026opencompassuniversalevaluationplatform}.
 }
 
\textbf{Other Interesting Aspects.}
Given the theoretical downgradation of causality, LLM infrastructures may explore intra-head parallelism during decoding, complementing existing inter-head parallelism approaches~\cite{cai2024medusa}, and relax strict token-level matching~\cite{jin2024ragcache,gim2024prompt} toward \emph{semantic approximation} during prefilling.
Speculatively, the similar Bayesian foundations of \model and PECJ~\cite{zeng2024pecj} may extend naturally toward a further joint optimization in AI and DB systems, breaking the long-standing modular barrier imposed by causality. 
%
In addition, they also link tightly with probalitics modelling, interpretability and AGI inspirations. Please refer to Appendix~\ref{sec:futurework} for more details.

\section{Conclusion}
\label{sec:conclusion}
We revisited the theoretical foundations of causality in language models from a Bayesian perspective.  
Through \model modelling, we showed that \emph{causal formulation is what you may downgrade if you look beyond data distribution and incorporate system behavior}. 
This is reaffirmed by a further validation and analysis regarding measure, control and evaluation of the latent variables. 
During the optimized control of approximation errors, a realizable and promising non-causal variational approximation prototype emerges as \family.
Beyond these immediate findings, this work calls for a broader shift 
from fitting data distributions under causal chains to explicitly characterizing system behaviors themselves as Bayesian latent variables (Definition~\ref{def:system_behavior}).  
We believe that modelling system behavior offers a unifying theoretical lens for understanding stability, efficiency, and adaptation in generative systems.

\newpage

\bibliographystyle{abbrv}
\bibliography{mylib,new_peckv}


\appendix
\section{More Discussions of Related Works}
\label{sec:appendix_other_related}
\subsection{Detailed Mismatches with Causality}
\textbf{Empirical Evidence.}  
It is show that even under strictly deterministic settings, model outputs can vary across runs~\cite{atil2024non}.  
Furthermore, future generation steps can retroactively influence past representations, such as embeddings and logits of generated tokens.  
This \emph{non-determinism} is typically framed in the literature as undesirable randomness or instability of runtime and hardware~\cite{atil2024non,song2024good,ouyang2025empirical}, and it is commonly treated as a bug to be corrected through numerical precision tuning, scheduling constraints, or infrastructure-level fixes.


\textbf{Theoretical Contradictions.}  
The \emph{Arrows of Time (AoT)}~\cite{pmlr-v235-papadopoulos24a} reveals a theoretical paradox that LLMs exhibit asymmetric perplexity when predicting the next token versus the previous one. This asymmetry contradicts the fundamental symmetry assumed in classical information theory~\cite{shannon1951prediction}. Proposed explanations, such as sparsity and computational asymmetry, are outside the scope of causal formulation itself and belong to algorithm aspects.

Another notable challenge arises from block diffusion~\cite{arriola2025block}. While still using causality, it offers a relaxation of the autoregressive (AR) causal chain. Instead of relying on token-by-token causality, block diffusion conducts a causal denoising process for each token block. Although it has not yet reached the scale of state-of-the-art AR models like ~\cite{yang2025qwen3,touvron2023llama,achiam2023gpt}, block diffusion has demonstrated improvements in global consistency and decoding speed.


\textbf{Implementation Strategies.}  
LLM inference typically consists of two phases, namely \emph{prefill} and \emph{decoding}. In both phases, recent strategies increasingly diverge from strict causality by involving new algorithm paradigms.
During prefill, causality implies that key-value (KV) cache reuse should follow strict prefix sharing. However, techniques like modular cache slicing~\cite{gim2024prompt,yao2025cacheblend} and prompt reconstruction~\cite{fan2024combining} allow relaxed reuse in specific domains, such as retrieval-augmented generation, thereby softening causal constraints.
In the decoding phase, techniques like speculative decoding~\cite{leviathan2023fast} and lookahead decoding~\cite{fu2024break} predict multiple tokens at once, which effectively breaking the causal step-by-step progression. More recently, multi-token prediction strategies~\cite{gloeckle2024better,cai2024medusa} have emerged, training auxiliary heads or modules to propose multiple candidate tokens per step for parallel decoding.

\subsection{Cognitive and Language Semantic Investigations}
\label{subsec:cog_discuss}
While \model highlights the irreducible impact of system behavior rather than relying solely on data-driven alignment, we are not the first to explore modelling beyond data distributions in the context of LLMs.  
Specifically, several recent studies report higher-order LLM behaviors that cannot be fully explained by data-distribution alignment, and most of them are linked with the language and cognitive capabilities themselves.
Examples include the emergence of metacognitive self-monitoring in mathematical reasoning~\cite{didolkarmetacognitive}, targeted representation interventions that boost arithmetic accuracy~\cite{zhang2024interpreting}, and the \emph{illusion of thinking} phenomenon, where apparent reasoning collapses under complexity shifts~\cite{illusion-of-thinking}.  
Other findings similarly highlight system-level tendencies: narrow fine-tuning on adversarial tasks can induce surprisingly broad misalignment effects~\cite{betley2025emergent}, and chain-of-thought reasoning can be decomposed into an information-theoretic process of incremental information gain, revealing hidden failure modes without explicit reasoning annotations~\cite{ton2025understanding}.  

We emphasize that it is not possible to exhaustively enumerate all such discoveries here, nor is our intention to challenge, rebuild, or extend any specific line of work.  
Instead, we regard these diverse findings collectively as a technique justification for modelling beyond data, as they point to persistent and structured tendencies in
the language part of system behavior that are not reducible to surface-level output patterns in token distributions. 
While \model is not intended to fully explain these phenomena, framing them together in future work may offer a principled way to study the deep mechanisms shaping LM behavior beyond token-level accuracy.  
Such an orthogonal perspective may eventually unify these seemingly disparate phenomena under a coherent dynamical framework.

\subsection{Modelling Behaviors and Strategies of LMs}
A growing body of work seeks to model or reverse-engineer how moren LM, especially LLMs, organize and deploy internal computational strategies.  
Among these behaviors, one prominent theme is \emph{emergent abilities}, where capabilities absent in smaller models appear abruptly at scale~\cite{zhao2023survey}, though some argue this arises from nonlinear evaluation metrics rather than genuine behavioral shifts~\cite{schaeffer2023emergent}.  
Recent studies broaden this view, showing that LLMs can spontaneously form multi-step contingency plans without explicit supervision~\cite{dong2025emergent}.  

Another line of work revisits \emph{in-context learning} (ICL), revealing multiple coexisting mechanisms, i.e., pure ICL, in-weights learning (IWL), and context-constrained IWL (CIWL) that compete and cooperate during training, with dynamic strategy switching explaining phenomena such as ICL transience~\cite{singh2023transient}.  

Beyond these training- or observation-centric analyses, \cite{wang2025unveiling} adopts a dynamical-systems lens, modelling inference as a discrete-time process and uncovering stable periodic patterns (attractor cycles) in output space across prompts, models, and decoding settings.  
Similarly, \cite{ji2025resist} introduces the concept of \emph{elasticity} in LLMs, showing that alignment effects are fragile and that models rebound toward pre-training distributions under small perturbations.  
Compression theory further links this resistance to Hooke’s law, suggesting that alignment effects are often superficial rather than deeply altering internal representations.

Our work is inspired by these insights but makes a major departure.  
Rather than treating system behaviors as byproducts of scale, data, or architecture~\cite{zhao2023survey,schaeffer2023emergent,dong2025emergent,singh2023transient,ji2025resist}, we posit that \emph{system behaviors themselves constitute part of the model}.  
This view is also more general and system-centric than~\cite{wang2025unveiling}, where the system serves primarily as an analytical metaphor rather than as a directly modelled entity.

\section{More Discussions of  \family}
\label{sec:more_exmaple}
This section is a more detailed version of Section~\ref{sec:so_what} with more derivations and justifications.
\subsection{Mean Field from Mathematical Physics}
\label{subsec:green_math_appendix}
\begin{equation}
q(\outputSeq, \sysBehavior) 
= \prod_{i \in \mathcal{I}} q_i(\text{variable}_i)
\label{eqn:mean_field}
\end{equation}

%
Mean-field formulations (Eq.~\ref{eqn:mean_field}) suggest that a family of tractable approximation mechanisms without assumed causality can in principle exist, enabling the conceptually free approximation of $q(i)$.
However, the notion of a ``mean field'' is itself overly broad and under-specified. 
In this paper, we restrict the variational family to one that has long been used to analyze complex electromagnetic and potential systems, i.e., the so-called \emph{Green-Shell}. 
In physics, Green-Shell methods~\cite{jackson2021classical,green1889essay} are widely used to approximate complex electromagnetic or potential fields. 
The basic idea is to replace a complicated continuous field with a set of concentric shells, each acting as an equipotential layer. 
Instead of tracking every fine-grained interaction, one only needs to compute how charges or sources are distributed across these shells. 
Compared to a preset causal chain, Green-Shells approximate a complex system through spatial partitioning rather than temporal dependency, making them inherently \emph{shared-nothing} and fully consistent with the assumptions of the mean-field family (Eq.~\ref{eqn:mean_field}) . 
Moreover, the number of shells can be scaled arbitrarily, directly paralleling how neural networks expand their capacity with more layers or parameters.
In the following, we demonstrate how this is applied to instantiate Eq.~\ref{eqn:mean_field} and formulate \family.

\textbf{Assumptions and Intuitions.}  
We begin by assuming that the output sequence $\outputSeq$ can be expressed by a hidden-state field $H$ defined with respect to a distance measure $r$ in a perturbative analysis, i.e., $\outputSeq \propto H(r)$.  
We do not restrict a concrete form of $r$ for two reasons:  
(1) distances in latent or embedding space can be freely projected or transformed by neural network mappings, and  
(2) $r$ is introduced only as a conceptual construct, and does not directly appear in the final loss formulation. Conceptually, the state space of $H$ can use tensors of either logits or embeddings, but in this paper we restrict to embeddings that typically involves less dimensions than logits, due to the notorious curse of dimensionality~\cite{oseledets2009breaking}. Consequently, $r$ is in the same dimension as embeddings.

\textbf{Conceptual Set-ups.}
After conceptually treating $\outputSeq$ as a field, we apply the following steps as what mathematical physics does.
First, the target field $H(r)$ of output $\outputSeq$ is divided into concentric shells as Eq.~\ref{eqn:appendix_green0}.
Each component is individual from the others, and can be defined as Eq.~\ref{eqn:appendix_green1}. $\rho(\mathbf{r}',\theta^{(i)})$ is the density function of the field, and it will be learned during model training with parameter space $\theta^i$ of layer $i$.  

For simplicity, we let the number of shells $T+1$ equal to the number of transformers used in the neural network, and each layer is one-one mapped with each layer.
\begin{align}
    \mathbf{H}(\mathbf{r}) &= \sum_{k=0}^{T} \mathbf{H}^{(k)}(\mathbf{r}) \label{eqn:appendix_green0} \\
    \mathbf{H}^{(k)}(\mathbf{r}) &= \int_{R_k < |\mathbf{r}'| < R_{k+1}} 
    \rho(\mathbf{r}',\theta^{(i)}) \frac{(\mathbf{r}-\mathbf{r}')}{|\mathbf{r}-\mathbf{r}'|^3} \, d^3\mathbf{r}' \label{eqn:appendix_green1}
\end{align}

Next, we extend the mean-field formulation in Eq.~\ref{eqn:mean_field} into a field-theoretic variational family (Eq.~\ref{eqn:appendix_green2}). 
In analogy with diffusion models, $\variationalProb(s^{i})$ conceptually conducts the denoising process, and is trained through the noising process $q^{noising}(s^i,Y)$.
However, unlike diffusion models, these processes are not anchored to the temporal order of $\inputSeq$ or $\outputSeq$; rather, they follow the intrinsic propagation of the field itself. 
In implementation, this corresponds to the neural network’s forward noising and backward denosing passes, which operate without any predefined causal chain. 
Concretely, in the forward noising step (Eq.~\ref{eqn:appendix_green3}), each layer perceives the inside of its shell as a ``clean system,'' while treating everything outside as Gaussian noise $\varepsilon^{(k)}$. 
This is analogous to a spherical cavity in electrostatics: the interior field seeks balance, while external disturbances only appear as noisy perturbations on the boundary.
In this sense, the \family instantiates a mean-field whose independence arises not from temporal factorization but from spatial partitioning.
\begin{align}
    q(Y,S) &   = \prod_{\text{For each layer } i} (\variationalProb(s^{i}) \times \variationalProb(y^i|s^{i}))\label{eqn:appendix_green2}
    \\
    q^{noising}(s^i,Y)  
    & \propto \sum_{k \leq i} \mathbf{H}^{(k)} + \sum_{k>i} \varepsilon^{(k)}\label{eqn:appendix_green3}
\end{align} 
\textbf{Rationales Behind.}
It is important to note that this formulation contains no explicit notion of input–output temporal causality, as the variational family is not tied to the order of $\inputSeq$ or $\outputSeq$. 
However, this does not reduce the model to a bag-of-words abstraction. 
By construction, the propagation of fields proceeds layer by layer, preserving an intrinsic \emph{directionality} analogous to temporal flow, even in the absence of an externally imposed sequence order. 
This directionality arises from the propagation of information through network depth and from the non-reversible dynamics of optimization and representation formation. 

Recent evidence further supports this interpretation. 
The \emph{Arrows of Time (AoT)} phenomenon~\cite{pmlr-v235-papadopoulos24a} demonstrates that LMs exhibit an inherent temporal asymmetry, as internal activations and attention flows encode a preferred direction consistent with information compression and entropy increase, even when explicit token order is removed. 
In this sense, directionality is more \emph{fundamental} than causality itself, as it arises not from pre-defined sequential rules, but from the irreversible flow of representation and optimization dynamics. 
\emph{Therefore, causality is not discarded but subsumed in \family}. What was previously interpreted as a left-to-right causal chain can instead be viewed as a manifestation of this emergent directional field.

\subsection{\family is Free From Additional UAT Residual Loss}
\label{subsec:free_justification}
As discussed in Section~\ref{sec:so_what}, \family involves (1) the temporal ordering constraint, (2) the spatial partition constraint, and (3) shared-nothing components equivalent to output $Y$.

The first property introduces no additional approximation error, as modern deep learning frameworks (e.g., PyTorch) execute layers in a strictly sequential forward pass. From the perspective of UAT, this temporal ordering does not impose any additional functional restriction that would require introducing an extra residual term beyond the intrinsic approximation error $\epsilon$ (Eq.~\ref{eq:uat-non}).
From a probabilistic graphical model perspective, the forward pass corresponds to a deterministic computation, implying that the event of satisfying the prescribed evaluation order has probability one under the induced execution measure.

The approximation error induced by the second constraint is also subsumed into the UAT error bound $\epsilon$. As a free approximation, each $\variationalProb(s^{i})$ does not constrain the UAT representational capacity in the infinite-width/depth limit, and is allowed to be \emph{sufficiently expressive}~\cite{hornik1989multilayer,bishop2006pattern} without restrictive structural assumptions (e.g., causal dependencies like $\variationalProb(s^{i}|s^{i-1})$) or additional fitting residuals (e.g., Bayesian hierarchy $|t_t(X)|$ in Eq.~\ref{eq:uat-causal}).
Under this regime, any discrepancy induced by the spatial partition constraint $p(s^i)$ can be absorbed through reweighting within ${\variationalProb(s^{i})}$~\cite{hornik1989multilayer,bishop2006pattern}. 

Furthermore, $q(y^i|s^i)$ ensures that approximation effects at each shell remain localized rather than being propagated through the coupled joint $q(Y|s^0, s^1, \ldots, s^T)$, leading to faster attenuation of prior-induced discrepancies under iterative updates~\cite{jordan1999introduction,turner2008two}, and thus typically faster convergence of model training.

Consequently, in the absence of structural restrictions on $\variationalProb(s^i)$, the overall approximation error bound remains determined by the intrinsic UAT error $\epsilon$, without an additional residual term such as $|t_t(X)|$.
\subsection{Mean Field into Training Loss}
\label{subsec:green_approx_appendix}
Eq.~\ref{eqn:appendix_field_loss} is a general training objective of arbitrary mean-field, 
where $p(H)$ denotes the true (unknown) distribution of hidden-state fields, and $q_\theta(H)$ is the variational family parameterized by $\theta$ in approximating the $H$.
For \family, Eq.~\ref{eqn:appendix_field_loss} naturally decomposes into layer-wise terms (Eq.~\ref{eqn:appendix_field_loss_layerwise}) given Eq.~\ref{eqn:appendix_green3}, and each $\mathcal{L}_{\text{field}}^{(i)}=\mathbb{E}_{p(H^{(i)})}\big[-\log q_\theta^{(i)}(H^{(i)})\big]$ is computed locally on shell $i$. 

\begin{equation}
    \mathcal{L}_{\text{field}} = \mathbb{E}_{p(H)}\big[-\log q_\theta(H)\big]
    \label{eqn:appendix_field_loss}
\end{equation}

\begin{equation}
    \mathcal{L}_{\text{field}} = \sum_{i=0}^{T} \mathbb{E}_{p(H^{(i)})}\big[-\log q_\theta^{(i)}(H^{(i)})\big]
    \label{eqn:appendix_field_loss_layerwise}
\end{equation}
This mirrors the way Green-Shells approximate a physical field by concentric equipotential layers in Mathematical Physics: every shell contributes its own loss, and the global approximation emerges from their superposition.  
In this sense, the variational objective preserves the structural independence of shells while assembling them into a coherent system-level approximation.

Nevertheless, a critical challenge arises because hidden-state fields and vocabulary tokens are not naturally aligned. 
This leaves two possible directions: one can \emph{discretize continuous fields into categorical indices}, or conversely, \emph{embed discrete symbols into continuous field formulations}.  
Both views have their own justification and difficulties.  
Discrete fields align neatly with vocabulary-level modelling and allow exact normalization via cross-entropy, but at the cost of discarding the continuous geometry of hidden states.  
Continuous fields preserve the geometric and dynamical structure observed in system trajectories, yet they face the challenge of defining or approximating a tractable normalization.  
Within the Green-Shell framework, however, this tension is secondary: \emph{whether discrete or continuous, the underlying task is mathematically a problem of field estimation.}  
Such dualities between continuous and discrete formulations are common in physics, where both perspectives may coexist and complement one another.  
In what follows (and in the Appendix), we briefly outline the discrete and continuous instantiations of this general objective.

\textbf{Discrete Establishing.}
If the hidden-state field $H^{(i)}$ is discretized into a finite vocabulary $\mathcal{V}$,
the variational distribution $q_\theta^{(i)}$ reduces to a categorical distribution (Eq.~\ref{eqn:appendix_discrete_general}), where $f_\theta^{(i)}(\cdot)$ is an arbitrary nonnegative scoring function.
\begin{equation}
    q_\theta^{(i)}(v) = \frac{f_\theta^{(i)}(v)}{\sum_{u \in \mathcal{V}} f_\theta^{(i)}(u)}, 
    \quad v \in \mathcal{V}
    \label{eqn:appendix_discrete_general}
\end{equation}
This introduces a nontrivial design space: different parametrizations of $f_\theta^{(i)}$ correspond to different ways of discretizing fields.  
For example, one may use  
\begin{itemize}
    \item Additive or multiplicative decompositions of hidden states. 
    \item Alternative normalizers beyond the softmax, such as sparsemax or entmax.
    \item Or structured mappings that reflect geometry in the field space.  
\end{itemize}

To make the above discussion concrete, let us consider a particularly simple instantiation.  
Suppose we take the scoring function in Eq.~\ref{eqn:appendix_discrete_general} to be an exponential of a linear logit,  
where $z_v^{(i)}$ is directly produced by the network.  
This choice is admittedly a form of “taking the easy way out”: it ignores possible geometric structure in the field, collapses all interactions into a single scalar score per token, and normalizes only through the exponential family.  
Substituting Eq.~\ref{eqn:appendix_exp_scoring} into Eq.~\ref{eqn:appendix_discrete_general} yields the familiar softmax distribution (Eqn~\ref{eqn:appendix_softmax_form}), and inserting Eq.~\ref{eqn:appendix_softmax_form} into Eq.~\ref{eqn:appendix_discrete_general} reduces the shell-wise objective to the cross-entropy as Eq.\ref{eqn:appendix_ce_loss}.
\begin{equation}
    f_\theta^{(i)}(v) = \exp(z_v^{(i)}), \label{eqn:appendix_exp_scoring}
\end{equation}

\begin{equation}
    q_\theta^{(i)}(v) = \frac{\exp(z_v^{(i)})}{\sum_{u \in \mathcal{V}} \exp(z_u^{(i)})}
    \label{eqn:appendix_softmax_form}
\end{equation}

\begin{equation}
    \mathcal{L}_{\text{field}}^{(i)} = - \sum_{v \in \mathcal{V}} p(H^{(i)}=v)\,\log q_\theta^{(i)}(v)
    \label{eqn:appendix_ce_loss}
\end{equation}

In other words, \emph{cross-entropy per layer is one particular instantiation within the broader field-estimation framework}, characterized by the exponential choice of $f_\theta^{(i)}$.
It survives largely as a convenient but ultimately limited option, and while richer discretization schemes may lead to more faithful field approximations, exploring them lies beyond the scope of this work and is left for future investigation. Recent works like~\cite{KalaiNachumVempalaZhang2025} also indicates the inherent suboptimal nature of cross-entropy from different aspects. 
Promising alternatives may involve non-softmax normalizers~\cite{peters2020sparse} or geometry-aware mappings~\cite{tang2024learning}, which suggest a broader design space than the exponential-softmax default.

\textbf{Continuous Establishing.}
Alternatively, one may retain the continuous nature of the hidden-state field $H^{(i)}$.  
In this view, the variational family $q_\theta^{(i)}$ is naturally expressed as a Gibbs distribution (Eq.~\ref{eqn:appendix_continuous_general}),  
where $\mathcal{H}_\theta^{(i)}(H)$ is an energy functional and $R_\theta^{(i)}$ is the intractable partition function.  
This formulation preserves the geometry of the field, but introduces a significant design space in how the energy $\mathcal{H}_\theta^{(i)}$ is chosen and how $R_\theta^{(i)}$ is approximated.  
\begin{equation}
    q_\theta^{(i)}(H) = \frac{\exp(-\mathcal{H}_\theta^{(i)}(H))}{R_\theta^{(i)}}
    \label{eqn:appendix_continuous_general}
\end{equation}

One particularly simple choice is to take $\mathcal{H}_\theta^{(i)}(H)$ as a quadratic energy like Eq.~\ref{eqn:appendix_quadratic_energy},
where $\hat{H}_\theta^{(i)}$ is the predicted clean field, and $R_\theta^{(i)}$ is simply assumed to be constant.

\begin{equation}
    \mathcal{H}_\theta^{(i)}(H) = \tfrac{1}{2}\|H - \hat{H}_\theta^{(i)}\|^2
    \label{eqn:appendix_quadratic_energy}
\end{equation}
Substituting Eq.~\ref{eqn:appendix_quadratic_energy} into Eq.~\ref{eqn:appendix_continuous_general} and evaluating the corresponding negative log-likelihood recovers a mean-squared-error(MSE) loss of Eq.~\ref{eqn:appendix_mse_loss}.
\begin{equation}
    \mathcal{L}_{\text{field}}^{(i)} \propto \|H - \hat{H}_\theta^{(i)}\|^2
    \label{eqn:appendix_mse_loss}
\end{equation}

Thus, {MSE can be understood not as an arbitrary engineering trick but as one particular instantiation of continuous field estimation, arising from the quadratic choice of energy.}  
This makes it numerically stable and convex, but also limited: the field is forced toward pointwise agreement, discarding global structural constraints and potentially collapsing nontrivial dynamics.  
More sophisticated energy functions, such as non-quadratic or learned functions~\cite{lecun2006tutorial,zhang2023persistently}, may yield richer and more faithful field approximations, though exploring them lies beyond the scope of this work.

\subsection{Detailed Implementation of \family}
\label{subsec:our_instantiation}
Although a full study of discrete (Eq.~\ref{eqn:appendix_discrete_general}) and continuous 
(Eq.~\ref{eqn:appendix_continuous_general}) Green-Shell approximation mechanisms are beyond the scope of this paper, 
a straightforward instantiation can be conducted as follows.

\textbf{Step 1: Continuous Field Constraint via Energy Functional.}
For each layer $i$, the hidden-state field $H^{(i)}$ is constrained by a field-theoretic 
energy functional. Inspired by the  \emph{Debye screening}~\cite{stern2007importance} in plasma physics, the Green kernel is screened in Eq.~\ref{eqn:appendix_kernel_layer}, 
and the corresponding field energy functional is given in Eq.~\ref{eqn:appendix_field_energy_layer}.  
\begin{equation}
    \mathcal{K}(r_{ij}, H^{(i)}_m, H^{(i)}_n) 
    = \frac{H^{-r_{ij}/\lambda}}{r_{ij}+e} 
    \Big(1-\gamma + \gamma \big(1 - \mathrm{sim}(H^{(i)}_m, H^{(i)}_n)\big)\Big)
    \label{eqn:appendix_kernel_layer}
\end{equation}

\begin{equation}
    \mathcal{L}_{\mathrm{field}}^{(i)}[H^{(i)}] 
    = \sum_{m,n} H^{(i)}_m \, \mathcal{K}(r_{mn}, H^{(i)}_m, H^{(i)}_n) \, H^{(i)}_n
    \label{eqn:appendix_field_energy_layer}
\end{equation}

Optimizing Eq.~\ref{eqn:appendix_field_energy_layer} plays a dual role. 
First, through the kernel definition in Eq.~\ref{eqn:appendix_kernel_layer}, the interaction between a predicted state $H^{(i)}_m$ and its target $H^{(i)}_n$ is not treated as independent noise but as mediated by a screened Green's function.  
Here $r_{ij}$ denotes the effective distance between prediction $i$ and target $j$, while $m,n$ index the corresponding token-level embeddings involved in this interaction.  
The exponential decay term ensures that long-range couplings are progressively attenuated, while the $1/r_{ij}$ singularity guarantees that no token position is ever completely isolated from the rest of the field.  
In addition, the similarity factor $\mathrm{sim}(H^{(i)}_m,H^{(i)}_n)$ discourages trivial collapse by explicitly penalizing pairs of tokens that converge to overly similar states.  
This introduces a diversity-promoting bias directly at the field-interaction level, with the strength of this effect controlled by the hyperparameter $\gamma \in [0,1]$.  
In contrast, the widely used MSE loss can be interpreted as a conceptually degenerate case of Eq.~\ref{eqn:appendix_field_energy_layer}, where the kernel reduces to an identity operator $ \mathcal{K}=\mathbf{I}$, thereby ignoring both spatial coupling and structural diversity, which is a convenient but ultimately impoverished approximation.  It is worth noting that $e$ is merely used for numerically preventing zero division, and we set it to $1\times10^{-4}$.

Second, Eq.~\ref{eqn:appendix_field_energy_layer} aggregates these pairwise interactions 
into a global quadratic form. This turns the hidden-state field $H^{(i)}$ into an 
energy configuration, where the total loss corresponds to the system’s field energy. 
Unlike local objectives, which only penalize pointwise deviations, this functional 
captures how deviations propagate and reinforce across the entire layer. As a result, 
the optimization enforces global coherence across shells: every unit must not only 
fit its local target but also remain consistent with the structure implied by 
its neighbors and the overall field.

Taken together, Eq.~\ref{eqn:appendix_kernel_layer} specifies the microscopic interaction law, 
and Eq.~\ref{eqn:appendix_field_energy_layer} evaluates the macroscopic energy of the system. 
By minimizing the latter, the network learns hidden-state fields that maintain 
internal consistency and diversity, without requiring any explicit causal ordering 
at the token level.

\textbf{Step 2: Black-Box Renormalization.}
The continuous field regulation of Eq.~\ref{eqn:appendix_field_energy_layer} ensures that each layer-wise hidden-state field 
$H^{(i)}$ is internally constrained by its own energy functional, so every layer 
``knows” what to optimize for. However, at the system level there remains the 
global partition function $R_\theta$ inherited from Eq.~\ref{eqn:appendix_continuous_general}. 
In physics, global normalizers such as $\int \exp(-\mathcal{H}(x))dx$ are notorious for 
the potential risk of unbounded energy~\cite{green1889essay}, such as the well-known 
\emph{ultraviolet catastrophes}~\cite{planck1900theory}, where infinitely many high-frequency modes contribute unbounded energy.  
Analogously, the partition function $R_\theta$ in Eq.~\ref{eqn:appendix_continuous_general} can be dominated by unlikely but high-entropy regions of the hidden-state space, destabilizing optimization.  
In LMs this would manifest as overemphasis on spurious rare or noisy token patterns, so we treat $R_\theta$ as a black box and replace it with the Black-Box Renormalization of Eq.~\ref{eqn:appendix_mix_target}, while leaving a more explicit and analytic renormalization to future work.

Concretely, we approximate the intractable $R_\theta$ by projecting fields back into token space through a mixed target distribution, inspired by label smoothing~\cite{szegedy2016rethinking}.  
Following Eq.~\ref{eqn:appendix_discrete_general}, we interpolate between the empirical dataset distribution $p_{\text{data}}$ and a uniform prior $u$, with mixing coefficient $\alpha \in [0,1]$:  
\begin{equation}
    q_{\mathrm{mix}}(v) = (1-\alpha)\, p_{\text{data}}(v) + \alpha \, u(v), \quad v \in \mathcal{V}
    \label{eqn:appendix_mix_target}
\end{equation}
This interpolation acts as a surrogate for explicit renormalization, preventing any single high-entropy mode from overwhelming training dynamics. It is worth noting that Eq.\ref{eqn:appendix_mix_target} also covers cross entropy as a special case by setting $\alpha$ to zero, and the tuning of $\alpha$ involves a trade-off between output diversity v.s. fitting accuracy.

The resulting black-box renormalization loss is then written as Eq.~\ref{eqn:appendix_renorm_loss}, where $q_\theta(v)$ denotes the model’s predicted distribution over the vocabulary.  
Eq.~\ref{eqn:appendix_renorm_loss} thus serves as a direct stand-in for the elusive $R_\theta$, enabling stable optimization without requiring its explicit form.  
\begin{equation}
    \mathcal{L}_{\mathrm{renormalization}} = - \sum_{v \in \mathcal{V}} q_{\mathrm{mix}}(v) \, \log q_\theta(v)
    \label{eqn:appendix_renorm_loss}
\end{equation}

Despite the absence of an explicit form of $R_\theta$ in Eq.~\ref{eqn:appendix_renorm_loss}, the intuition is straightforward.  
By interpolating between the empirical data distribution and a uniform prior, probability mass is redistributed so that no single mode dominates.  
As a result, Black-Box Renormalization provides a pragmatic closure to the variational family, ensuring that the continuous field formulation can be consistently reconciled with discrete token alignment.  

\textbf{Putting Things Together.}  
To approximate a mixture of discrete and continuous fields, our standard instantiation integrates two complementary steps.  
First, each layer is parameterized by a continuous \emph{field energy functional} (Eq.~\ref{eqn:appendix_field_energy_layer}), which preserves local structure and prevents collapse into bag-of-words abstractions.  
Second, because the global partition function $R_\theta$ (Eq.~\ref{eqn:appendix_continuous_general}) is intractable and may diverge, we employ a \emph{Black-Box Renormalization} (Eq.~\ref{eqn:appendix_renorm_loss}) to project the system back into discrete token space.  

The renormalization can be applied in two regimes.  
When restricted to the final \texttt{lm\_head} or merely the last several layers, it corresponds precisely to the backpropagation-free architecture introduced by~\cite{li2025noprop}.  
This setting is computationally efficient but may retain residual inconsistencies.  
Alternatively, applying renormalization across the entire network yields more consistent yet costlier alignment.  
We treat the choice between these two regimes as an empirical question and evaluate both in Figure~\ref{fig:bonus}; a more complete investigation is left for future work.

\textbf{Possible Variants.}  
While the two-step procedure follows the physical intuition of approximating fields, it is not the only viable formulation.  
In practice, one may \emph{omit the first step}, allowing the entire network to be parameterized by the renormalization term.  
This aligns better with standard training pipelines and offers direct optimization toward target texts.  
However, doing so disables the backpropagation-free possibility of \family, since gradients from renormalization must now traverse all layers without a continuous-field parameterization.  
Conversely, one may \emph{omit the second step} and allow the continuous field alone to drive self-consistency.  
Although this reduces data dependency across layers and simplifies training, it risks divergence by letting $R_\theta$ grow without bound.  

\textbf{Block-wise Field Approximation.} We follow the block-wise generation of block diffusion. 
 We retain causal masks between blocks, but replace the standard temporal denoising process inside each block with non-causal \family approximation.
{
Different from block diffusion, an \emph{interleaved reconstruction} replaces the noise schedule in the training phase. Masked blocks (e.g., $B$ and $D$ in $A[\text{noise}]C[\text{noise}]$) are trained to reconstruct using surrounding unmasked blocks as boundary conditions without any noising/denoising chain, therefore shifting the objective from following a preset noise decay schedule to directly controlling \(\sysBehavior\) within each block. 
The inference phase is similar to block diffusion, where we append a pure noise block to the current context embedding, forward it through the non-causal Green Shells, and use the decoded tokens to update the prompt for the next block’s generation.}

\section{\snr as a Proxy for UAT Error Bound}
\label{sec:snr_detailes}
\textbf{Lazy Training Background}.
In the regime of \textit{lazy training}~\cite{jacot2018neural},  parameter updates remain within the linearization region 
of the initialization, suppressing large-scale representation drift. 
Such a regime isolates the \emph{system behavior} induced by the model architecture itself 
from data-driven feature learning dynamics, 
thereby providing a cleaner environment for examining model capabilities and error bounds.

Under the lazy training  regime, the evolution of a neural network $F(X)$ can be linearized around its initial parameters $\theta_0$. The approximation capability is governed by the Empirical NTK matrix $\mathbf{K} \in \mathbb{R}^{N \times N}$, where $N$ is the number of samples when obtaining the NTK matrix. Following~\cite{jacot2018neural}, we define the spectral decomposition of the Gram matrix as Eq. ~\ref{eq:spectral_decomp}.
\begin{equation}
\mathbf{K} = \sum_{i=1}^{N} \lambda_i \mathbf{v}_i \mathbf{v}_i^T
\label{eq:spectral_decomp}
\end{equation}

\textbf{Signal, Noise and Bounds}.
Following Random Matrix Theory (RMT) and the Marchenko-Pastur law~\cite{martin2021implicit}, we further partition the spectrum into a \textit{Signal Subspace} and a \textit{Noise Bulk} at the median index $m = N/2$, and the power of each component is given by Eq.~\ref{eq:power_defs}.
Note that, the optimal division threshold should be 
 $4/\sqrt {3}$~\cite{gavish2014optimal}, but here we use $2$ for implementation simplicity.
With the decomposed signal and noise, we define Signal-to-Noise of NTK (\snr for short) as Eq.~\ref{eq:snr_def}.
\begin{equation}
P_{sig} = \sum_{i=1}^{m} \lambda_i, \quad P_{noise} = \sum_{i=m+1}^{N} \lambda_i
\label{eq:power_defs}
\end{equation}
\begin{equation}
\text{SNR} = \frac{P_{sig}}{P_{noise}}
\label{eq:snr_def}
\end{equation}

Following the localized Rademacher complexity theory~\cite{bartlett2002rademacher,caponnetto2007optimal}, the UAT error $\epsilon$ of Eq.~\ref{eq:uat-non} or ~\ref{eq:uat-causal} is related to the complexity of the hypothesis class restricted to the functions with low variance. In the NTK regime, such complexity that leads to $\epsilon$ is dominated by the noise bulk $P_{noise}$ relative to the stabilizing signal power $P_{sig}$. By normalizing the spectral leakage at $P_{noise}$ against the empirical kernel scale at $P_{sig}$, we obtain Eq.\ref{eq:noise_governed_bound}.
$C$ is irrelevant to data, but only affected by the neural network architectural and the complexity of task.
\begin{equation}
\epsilon \propto C \cdot \sqrt{\frac{P_{noise}}{N \cdot P_{sig}}}
\label{eq:noise_governed_bound}
\end{equation}
\begin{equation}
\epsilon \propto \sqrt{\frac{1}{\text{SNR}}} 
\label{eq:final_proxy}
\end{equation}

By keeping (1) the same neural network, (2) the same fitting object, and (3) the same number of NTK samples $N$, Eq.~\ref{eq:noise_governed_bound} and ~\ref{eq:snr_def} leads to Eq.~\ref{eq:final_proxy}. Thus, the \snr calculated via Eq.~\ref{eq:snr_def} serves as a rigorous proxy for the structural stability and approximation convergence of $f$ under the UAT framework.

\section{Other Experimental Settings}
\label{sec:setups}
\paragraph{Overall Settings.}
Our experiments are deliberately conducted under a \textbf{lazy-training regime}~\cite{jacot2018neural}.
This configuration ensures that parameter updates remain within the linearization region 
of the initialization, suppressing large-scale representation drift. 
Such a regime isolates the \emph{system behavior} induced by the model architecture itself 
from data-driven feature learning dynamics, 
thereby providing a cleaner environment for examining the theoretical properties 
of the proposed non-causal variational family \family and \model itself.

To highlight the structural implications of our proposed variational family \family, 
we conduct a controlled comparison against two canonical baselines: 
(1) the textbook-level \textbf{autoregressive (AR)} model, representing strict token-level causality, 
and (2) the recent \textbf{block diffusion} mode~\cite{arriola2025block}, which realizes a Markovian causal denoising process. 
All models are trained from scratch under identical configurations, 
using the same \textbf{150M-parameter backbone} with \textbf{ROPE positional encoding} 
and the same optimizer settings. 
Each transformer layer is implemented as a simplified Qwen-3 transformer~\cite{yang2025qwen3} without SwiGLU activations or MoE routing,  
so that the purity of the dynamics remains unaffected by optimization-specific modifications.  
We utilize the original tokenizer and embeddings of Qwen-3 for simplicity, and let the hidden states sized at $2560-D$.
This design choice makes the subsequent analysis of NTK spectrum and field interactions directly comparable across variants, without interference from activation scaling or expert gating effects.
For fairness, both \family and block diffusion share an identical \textbf{block mask} 
and perform \textbf{parallel decoding over 5 tokens per step}, 
whereas the AR baseline decodes one token at a time. 

Conceptually, these three models form a progressively relaxed causal hierarchy. 
The AR model enforces strict autoregressive dependence across tokens; 
the block diffusion model relaxes this by introducing a Markov chain in latent space, 
retaining causality only across blocks but not individual tokens; 
and our \family goes one step further by \emph{removing the causality within each block}. 
Instead, \family relies solely on the intrinsic directionality of the latent field 
and the spatial inductive bias introduced by ROPE to preserve contextual coherence. 
This design embodies a causality-free generative principle consistent 
with our theoretical framework of system behavior modelling.
We construct a 10k passage subset from the dataset (\texttt{TinyStories}~\cite{eldan2023tinystoriessmalllanguagemodels} by default) as training data, and each contains about $100$ tokens.  
For single-objective settings of AR and block diffusion, we train for 4 epochs, while for combined objectives involved in \family (Appendix~\ref{subsec:our_instantiation}), each component is trained for 2 epochs by default (total 4). The batch size is by default set to $32$, and general tendencies reserve if reduced to 8 or 16.

We utilize the Adam optimizer with a fixed learning rate of $1\times10^{-4}$, rather than AdamW or any adaptive scheduling variant. 
The reason is that the lazy-training regime assumes parameter updates remain within an infinitesimal neighborhood of initialization, such that the model behaves as a first-order kernel expansion.  
In this regime, introducing additional scheduling effects, such as weight decay or adaptive learning-rate modulation in AdamW, effectively alters the curvature of the optimization trajectory and changes the implicit NTK geometry.  
These adjustments can accelerate convergence but break the fundamental assumption of parameter stagnation ($\|\Delta\theta\|\!\ll\!\|\theta_0\|$), thereby violating the conditions required for theoretical correspondence between training dynamics and kernel evolution.  
By fixing the learning rate and removing extraneous scheduling, we preserve a \emph{faithful lazy-training regime} in which observed differences in convergence and stability can be attributed solely to model structure rather than optimizer dynamics.
For the specific hyperparameters of \family, we set $\alpha = 0.25$ and $\gamma = 0.5$, which serve as balancing coefficients between reconstruction accuracy and output diversity. 
While this configuration may not be optimal across all conditions, it was empirically validated as stable and effective within our evaluation scope.  
From a theoretical standpoint, $\alpha$ and $\gamma$ can be interpreted as modulating the relative strength of self- and cross-interaction terms.  
Future work may consider learning-based adaptation of these coefficients or, more fundamentally, analytical derivations that allow them to be absorbed into the renormalized kernel, thereby removing manual tuning altogether.

\paragraph{Machines.}
Our evaluations have been conducted on NVIDIA RTX~4090, A6000~Ada, and AMD RX~9070~XT GPUs, and a single card is sufficient. 
For most evaluations except Figure~\ref{fig:cross_correlation}, a minimum of 16\,GB VRAM is required (with batch size $8\sim16$), while 24\,GB is recommended (with batch size $32$).
For Figure~\ref{fig:cross_correlation}, 32\,GB VRAM is mandatory in some LLMs.

\textbf{Relative Error at the Same Random Seed}.
Given the same random seed, it is the runtime and hardware behavior that involves minor drifts, therefore, the relative error is evaluated to be strictly bounded.
The relative error across repeated runs on the same machine is within $\pm4\%$, while the discrepancy across different NVIDIA CUDA machines is within $\pm7\%$.
The performance is comparable across CUDA and ROCm platforms, and relative error is bounded within $\pm12\%$.

\begin{figure}[t]
\centering
     \begin{minipage}[c]{0.75\textwidth}
     \subfigure[Error bar for \snr.]{
            \includegraphics[width=0.4\textwidth]{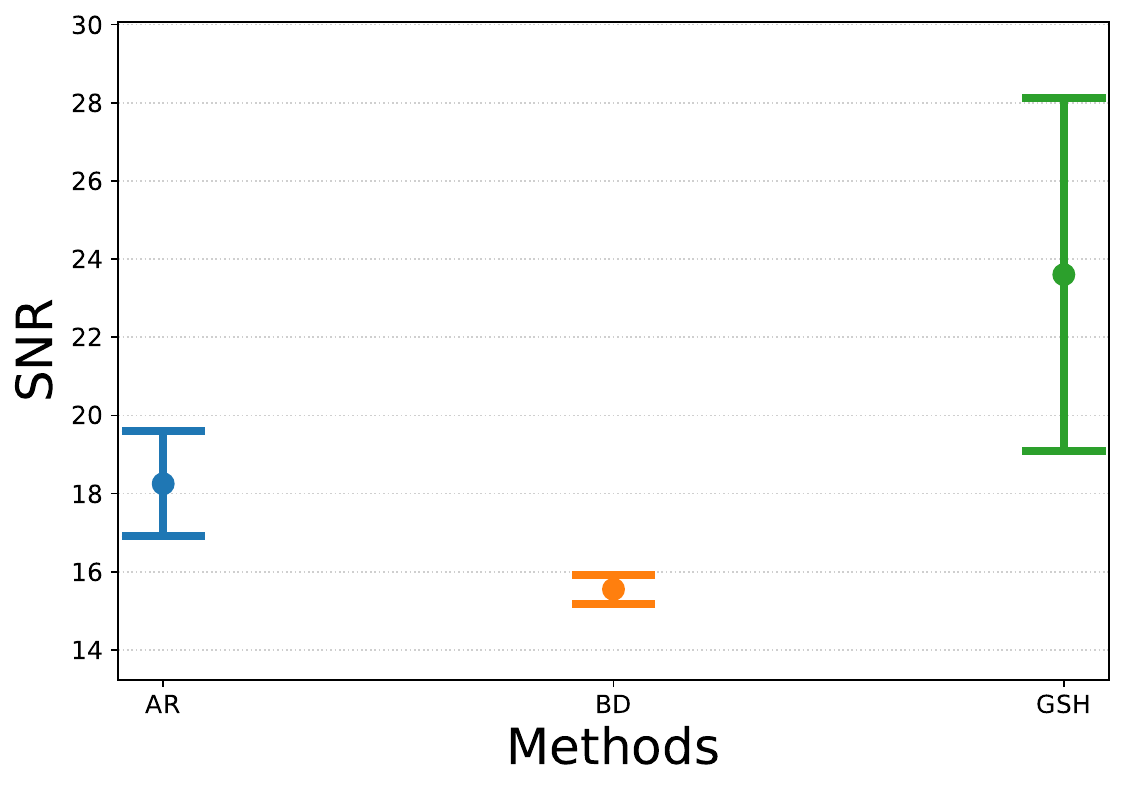}
            \label{fig:error_bar_snr}
        }
        \subfigure[Error bar for \mlc. ]{
            \includegraphics[width=0.4\textwidth]{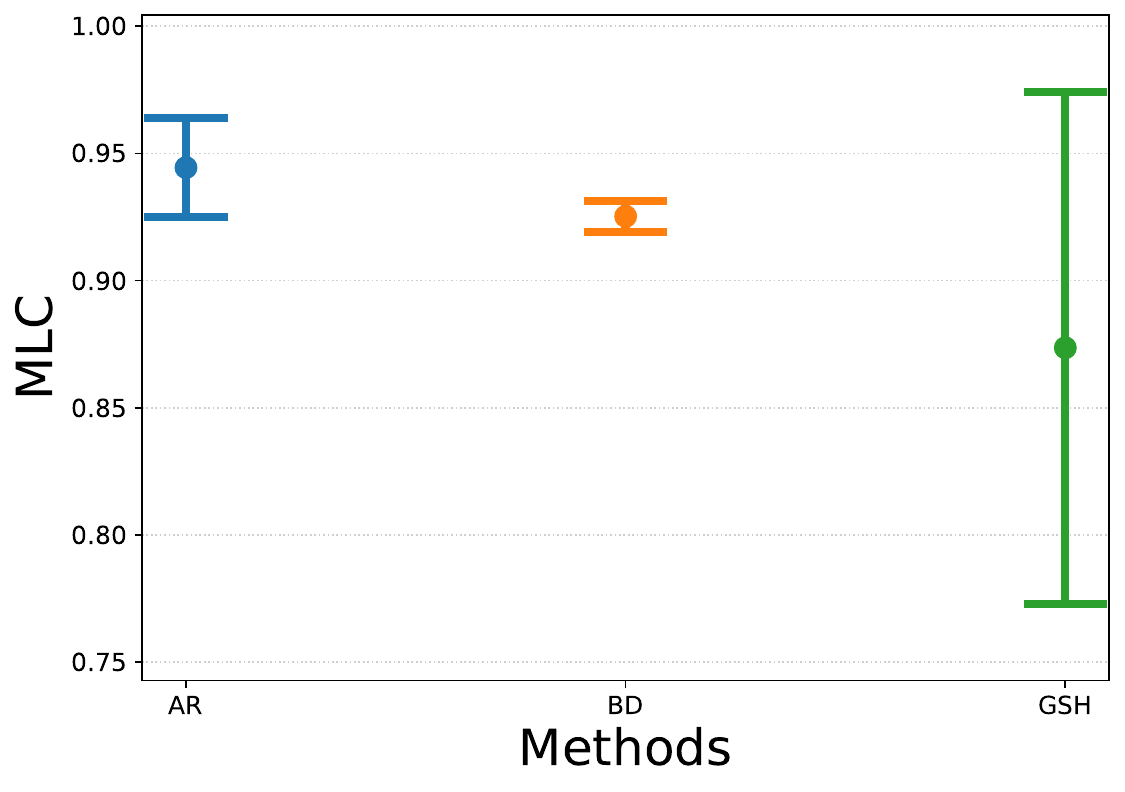}
            \label{fig:error_bar_mlc}
        }
\caption{Error bar across $50$ different random seeds.}
    \label{fig:error_bar}
    \end{minipage}
  \end{figure}

\textbf{Error Bar at the Different Random Seeds}.
Different random seeds involve different optimizer initialization, while it does not affect the overall tendencies, the detailed metric value is affected. We therefore report the $1\sigma$ error bars across $50$ random seeds on TinyStories datasets as example (Figure~\ref{fig:error_bar}), and other cases remain similar.

  \begin{figure}[t]
  \begin{minipage}[c]{0.65\textwidth}
        \subfigure[SNR]{
            \includegraphics[width=0.46\textwidth]{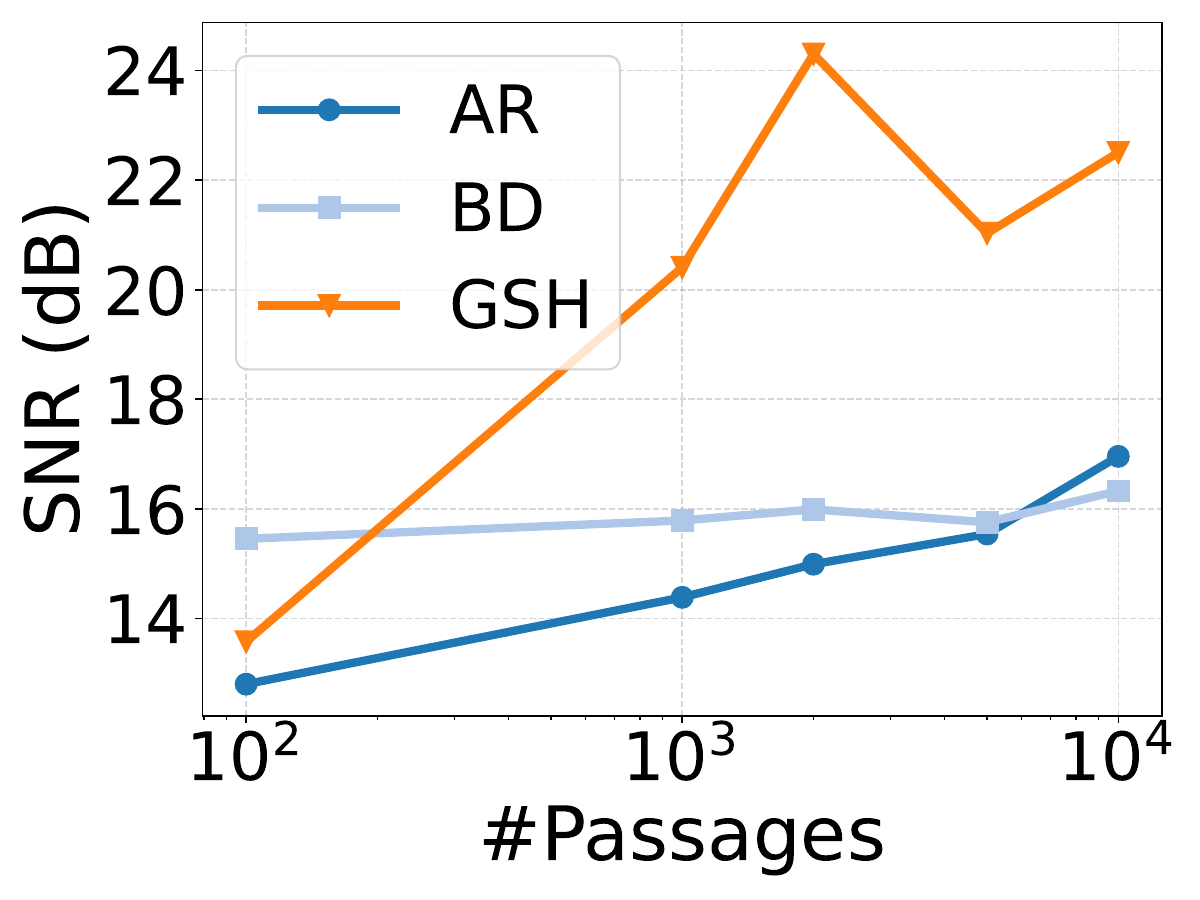}
        }
          \subfigure[MLC]{
            \includegraphics[width=0.46\textwidth]{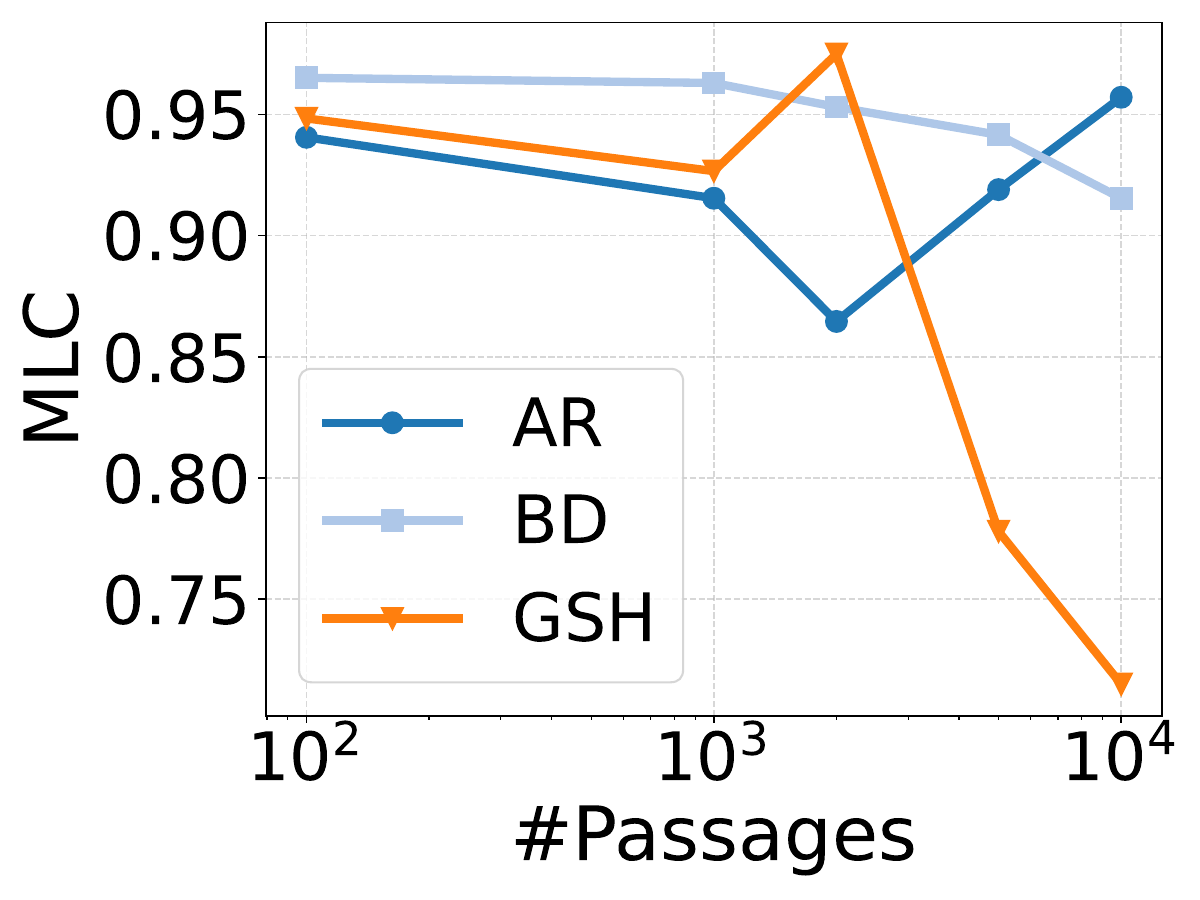}
        }
        \caption{Data scalability.}
    \label{fig:data_scalability}
    \end{minipage}
      \begin{minipage}[c]{0.32\textwidth}
      \centering
        \includegraphics*[width=0.99\textwidth]{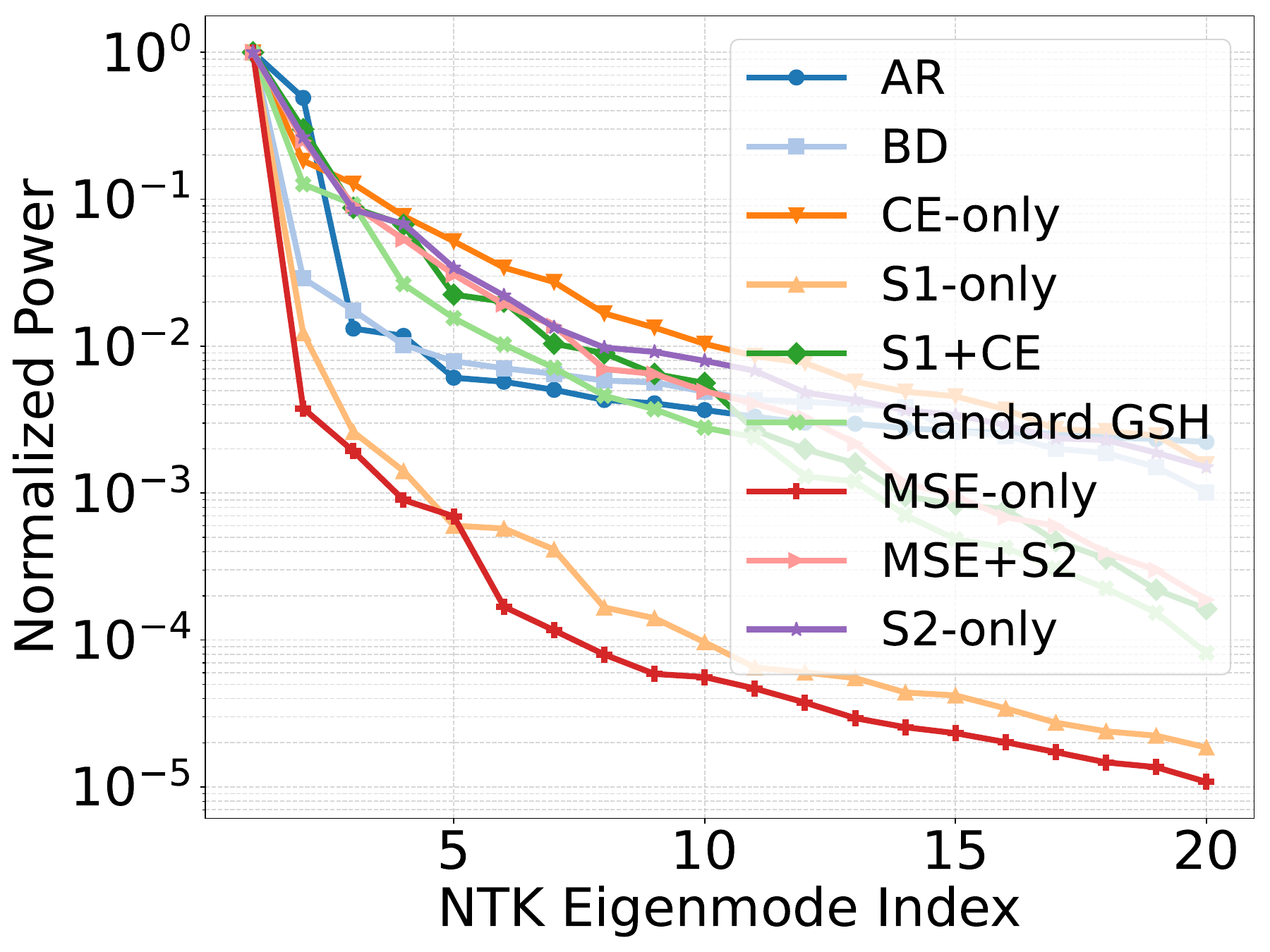}
        \caption{NTK spectra of ablation study.}  
    \label{fig:ablation}
    \end{minipage}
      \end{figure}
Throughout this paper, we report results obtained on the RTX~4090 for consistency.
\section{Additional Experimental Studies}
\label{sec:exp_additional}
We report the additional sensitivity and ablation studies regarding \family in this Section.
\subsection{Additional Sensitivity Study}
In this part, we take \texttt{TinyStory} as the example dataset while keeping other settings default to Table~\ref{tab:e2e}, varying the training data scale from $100$ to $10{,}000$, and each passage involves $100$ tokens.  The resulting \snr and \mlc are shown in Figure~\ref{fig:data_scalability}.
We deliberately limit the data scale such that the number of tokens remains \emph{far smaller} than the number of parameters. Beyond this point, the training dynamics would exit the lazy training regime, rendering the NTK spectral analysis no longer a reliable proxy for the theoretical UAT error bound.
Within the full examined range, \family consistently achieves the highest \snr, and the overall growth speed is also highest.
The observed sharp curve around $2000$ is primarily attributed to the current proof-of-concept implementation, particularly the heuristic renormalization and unoptimized hyperparameters that have not been fully resolved (Section~\ref{sec:so_what} and Appendix~\ref{sec:more_exmaple}).
More importantly, both \family and BD exhibit an overall decreasing trend in \mlc as data scale increases, 
while AR baseline shows the opposite trend. This suggests that \family and BD have a weaker tendency toward overfitting compared to the standard causal formulation as Bayesian approximations. Furthermore, \family is even better due to the greater relaxation of causality.
\subsection{Additional Ablation Study}
\label{sec:ablation}
In this part, we explore how discrete and continuous establishing act themselves and interact with each other. Specifically, we conduct an ablation study by comparing \textbf{standard} \family (Section~\ref{subsec:our_instantiation}) with the following approaches.
\begin{enumerate}
  \item \textbf{MSE+S2}: the Green-kernel parameterization (Eq.~\ref{eqn:appendix_field_energy_layer}) of standard setting in the first step is simplified by MSE.
  \item \textbf{S1+CE}: the renormalization part (Eq.~\ref{eqn:appendix_mix_target}) in the second part is reduced into cross entropy renormalization.  
  \item \textbf{S1-only}: the discrete setup is removed, and there is only 10 epochs of Green-kernel-based continuous training.  
  \item \textbf{S2-only}:  the continuous setup is removed, and there is only 10 epochs of mixed-target-based discrete renormalization. 
  \item   \textbf{MSE-only}: similar to \textbf{S1-only}, but utilize MSE.
  \item \textbf{CE-only}: similar to \textbf{S2-only}, but utilize cross entropy.
\end{enumerate}
We keep the basic setup the same as Figure~\ref{fig:ntk_e2e}(a) and report the resulting NTK spectra in Figure~\ref{fig:ablation}. There are two primary takeaways.

First, both continuous setup and discrete renormalization are required in \family. When everything is discrete (\textbf{S1-only} or \textbf{CE-only}), there are even richer peripheral features learned with $MLC < 0.7$. However, the suppression of the noise bulk is even weaker than taht in AR or BD, making $SNR < 16$ dB. On the other hand, when everything is continuous (\textbf{S2-only} or \textbf{MSE-only}), there is an extremely high noise suppression with $SNR > 30 dB$, but most NTK energy is concentrated with $MLC > 0.95$. This is because both continuous fields and discrete tokens are important, and one has to align them during the parameterization (Appendix~\ref{subsec:green_math_appendix}).

Second, the spectra also indicate an interestingly complex \textit{field-interaction effect} between continuous and discrete components. While the noise suppression of \textbf{S1-only} is weaker than that of \textbf{MSE-only} (i.e., 34.1 dB vs. 36.2 dB in SNR), the results become opposite when combined with continuous renormalization. Specifically, standard \family leads to higher SNR than \textbf{MSE+S2} (22.5 dB vs. 20.3 dB).
In continuous–discrete coupled physics problems (e.g., hybrid particle–field simulations~\citep{birdsall2018plasma}), the two fields are \emph{not linearly separable}, and we suspect a similar property holds for language modeling.  
\section{Detailed Discussions of Future Works}
\label{sec:futurework}
\paragraph{Training New Foundation Model Families.} 
While our study focused on the theoretical formulation of \model and its valid instantiation \family under the lazy-training regime, 
Proposition~\ref{pro:decouple_elbo} and Theorem~\ref{thm:elbo} alone with their empirical validations at Section~\ref{sec:exp} point naturally toward a broader class of \emph{non-causal foundation models} 
whose inductive biases align with global system behavior rather than local data distributions. Specifically, future works may inherit \family's possibility of 
backward-propogation-free training~\cite{li2025noprop} at large scale 
, while further enrich or extend the 
(1) system-behavioral modeling, (2) generalized continuous and discrete loss functions and (3) field-theoretic optimizations.
These directions follow directly from our mathematical derivations and may provide a theoretical foundation for more efficient, stable and adaptive large-model families. 

\paragraph{LLM Inference Infrastructure.} 
Most existing acceleration strategies~\cite{jin2024ragcache,gim2024prompt,fan2024combining,gloeckle2024better,cai2024medusa} 
treat inference efficiency as an engineering problem, decoupled from the intrinsic system behavior of models. 
However, since causality has been shown to be unnecessary and even suboptimal, 
new architectural pathways become possible. 
For decoding acceleration, intra-head parallelism may enable a single head to emit multiple tokens through field reconstruction, as in \family, complementing existing inter-head parallelism approaches~\cite{cai2024medusa}. 
For prefilling, strict token-level matching~\cite{jin2024ragcache,gim2024prompt} could be relaxed toward \emph{semantic approximation}, 
allowing more flexible reuse of intermediate states once causal constraints are lifted.

\paragraph{Database (DB) for AI and AI4DB.}  
As surveyed in~\cite{9094012}, \emph{AI4DB} often treats models as black-box replacements for components such as learned indexes or query optimizers, 
while \emph{DB4AI} applies database principles like indexing or materialized views to isolated ML operators.  
Both remain modular and loosely coupled, limiting deep integration between data management and model dynamics.  
When AI is instantiated as an LLM, the causality assumption becomes a major obstacle: 
data reuse tends to be all-or-nothing, requiring either full re-computation or full replay of token sequences.  

However, by explicitly modeling \emph{system behavior}, \model downgrades causality into a suboptimal special case of LLM modeling and enables us to overcome it, potentially allowing data flow and the LLM to be co-optimized as interacting processes rather than sequential stages.  
Under this view, formulations beyond \family may extend naturally toward a \emph{field-theoretic integration} of DB and AI systems, 
where query engines and models are represented as a unified system behavioral field.  
Such a perspective suggests that storage, retrieval, and inference could become continuous operations of the same underlying field of a system, 
breaking the long-standing modular barrier imposed by causality.

\paragraph{Probabilistic Modeling.}
We conjecture that our perspective and AoT (as discussed in Section~\ref{sec:related}) may have a deeper connection rather than orthogonal discoveries of abnormalies. AoT attributes directional asymmetry to structural and computational properties of the model, while our framework integrates such properties as manifestations of a latent \emph{system behavior} set $\sysBehavior$.
Viewed through this lens, AoT could be interpreted as probing a specific directional derivative of system behavior with respect to token order, which is an observable imprint of internal dynamics that causal modeling alone cannot capture.
This suggests a potential unifying research direction: understanding how 
$\sysBehavior$ interacts with generation order, and whether AoT-like asymmetries are surface expressions of a more general system-level probabilistic structure. Exploring this possibility may offer a principled bridge between computational asymmetry and non-causal variational modeling, and we consider this a promising avenue for future work.

\paragraph{Interpretability and AGI Inspirations.}  
\model also connects to recent studies on interpretability~\cite{illusion-of-thinking,ton2025understanding,KalaiNachumVempalaZhang2025}: 
phenomena such as hallucinations and repetitions may be understood not as isolated generation errors, but as manifestations of system behavior.  
This view enables explanations grounded in Bayesian principles rather than post-hoc probing or mechanistic attribution.  

Speculatively, we observe tentative parallels to psychological studies on grandiose narcissism and narcissistic personality disorder~\cite{zajenkowski2021telling,weinberg2022narcissistic}, 
which suggest that humans may likewise exhibit persistently confident yet erroneous responses, in analogous to LLMs operating under misaligned field configurations (the \textbf{S1-Only} or \textbf{MSE-Only} setting in Figure~\ref{fig:ablation}).  
This raises a natural opening question: \emph{is the system behavior $\sysBehavior$ in Proposition~\ref{pro:decouple_elbo} a controllable, dominant component of intelligence across domains?}

\section{Potential Risks of Misuse}
\label{sec:risks}
Given the disruptive nature of rethinking causality and highlighting system behavior in LMs, it is important to acknowledge potential misuse risks as follows alongside scientific contributions.

First, malicious actors could intentionally exploit \model to produce deceptive outputs, create self-reinforcing dependency loops, or embed covert triggers, underscoring the dual-use potential of this framework.  
Because our formulation makes system behavior explicit, it may expose new attack surfaces analogous to energy-landscape manipulations of model weights, adversarial training vulnerabilities, or even malicious infrastructure components.

Second, although our work demonstrates constructive uses of mathematical-physics field formulations in \family, the same principles could be misused.  
Adversarial or pathological renormalizations might be engineered to induce misleading behaviors even on well-curated datasets, revealing potential vulnerabilities in model trustworthiness.  

Third, 
our initial attempts of misconfigurations
indicates tentative parallels between incorrect LM configurations and pathological human traits such as narcissistic personality disorder~\cite{weinberg2022narcissistic}.  
Although still speculative now, this analogy raises concern about systems that appear aligned at the task level yet exhibit unstable or manipulative internal dynamics. This may lead to risks conceptually related to emergent misalignment~\cite{betley2025emergent}.  

At the same time, open publication may bring substantial positive impact.  
By articulating this framework explicitly, we enable the community to anticipate vulnerabilities, design countermeasures, and foster interdisciplinary inquiry into alignment, robustness, and the nature of intelligence itself.  
Moreover, analogies to pathological fields may inspire new diagnostics for detecting early signs of degeneracy in model behavior.  
We therefore believe that the benefits of transparency and discussion outweigh the associated risks.

\newpage

\end{document}